\documentclass{article} % For LaTeX2e
\usepackage{nips15submit_e,times}
\usepackage{hyperref}
\usepackage{url}
\usepackage{subfigure}
\usepackage{amsmath}
\usepackage{amssymb}
\usepackage{soul}

\usepackage{stmaryrd}
\usepackage{bbm}
\usepackage{cite}
\usepackage{url}
\usepackage[numbers]{natbib}
\usepackage{algorithm}
\usepackage{algorithmic}
\usepackage{graphicx}

\title{Combinatorial Bandits Revisited}

\author{
Richard Combes$^*$  \qquad M. Sadegh Talebi$^\dagger$ \;\qquad Alexandre Proutiere$^\dagger$ \qquad Marc Lelarge$\ddag$ \\
$^*$ Centrale-Supelec, L2S, Gif-sur-Yvette, FRANCE \\
$\dagger$ Department of Automatic Control, KTH, Stockholm, SWEDEN \\
$\ddag$ INRIA \& ENS, Paris, FRANCE \\
\texttt{richard.combes@supelec.fr, \{mstms,alepro\}@kth.se, marc.lelarge@ens.fr}
}

\newtheorem{theorem}{Theorem}

\newtheorem{lemma}{Lemma}
\newtheorem{proposition}{Proposition}
\newtheorem{corollary}{Corollary}

\def\EE{{\mathbb E}}
\def\PP{{\mathbb P}}
\def\RR{{\mathbb R}}

\def\Pcal{{\mathcal P}}

\def\Ocal{{\mathcal O}}

\def\Hcal{{\mathcal H}}

\def\Mcal{{\mathcal M}}

\def\Kcal{{\mathcal K}}

\def\Zcal{{\mathcal Z}}

\def\kl{\mathrm{KL}}
\def\klber{\mathrm{kl}}

\newcommand{\bp}{\noindent{\bf Proof.}\ }
\newcommand{\ep}{\hfill $\Box$}
\newcommand{\BEAS}{\begin{eqnarray*}}
\newcommand{\EEAS}{\end{eqnarray*}}
\newcommand{\BEA}{\begin{eqnarray}}
\newcommand{\EEA}{\end{eqnarray}}
\newcommand{\BEQ}{\begin{equation}}
\newcommand{\EEQ}{\end{equation}}
\newcommand{\BIT}{\begin{itemize}}
\newcommand{\EIT}{\end{itemize}}
\newcommand{\BNUM}{\begin{enumerate}}
\newcommand{\ENUM}{\end{enumerate}}
\newcommand{\al}[1]{ \begin{align} #1  \end{align}}

\newcommand{\als}[1]{ \begin{align*} #1  \end{align*}}
\newcommand{\eqs}[1]{ \begin{equation*} #1  \end{equation*}}
\newcommand{\sk}{\nonumber\\}

\newcommand{\el}{\end{flushleft}}
\newcommand{\bl}{\begin{flushleft}}

\newcommand{\NN}{\mathbb{N}}
\newcommand{\indic}{\mathbbm{1} }

\newcommand{\OPT}{V}
\newcommand{\tto}[1]{\underset{#1}{\to}}

\newcommand{\algos}{\textsc{ESCB}}

\nipsfinalcopy % Uncomment for camera-ready version

\begin{document}

\maketitle

\begin{abstract}
This paper investigates stochastic and adversarial combinatorial multi-armed bandit problems. In the stochastic setting under semi-bandit feedback, we derive a problem-specific regret lower bound, and discuss its scaling with the dimension of the decision space. We propose \algos, an algorithm that efficiently exploits the structure of the problem and provide a finite-time analysis of its regret. \algos~ has better performance guarantees than existing algorithms, and significantly outperforms these algorithms in practice. In the adversarial setting under bandit feedback, we propose \textsc{CombEXP}, an algorithm with the same regret scaling as state-of-the-art algorithms, but with lower computational complexity for some combinatorial problems.
\end{abstract}

\section{Introduction}
\label{sec:intro}

Multi-Armed Bandits (MAB) problems \cite{robbins1985some} constitute the most fundamental sequential decision problems with an exploration vs. exploitation
trade-off. In such problems, the decision maker selects an arm in each round, and observes a realization of the corresponding unknown reward distribution. Each decision is based on past decisions and observed rewards. The objective is to maximize the expected cumulative reward over some time horizon by balancing exploitation (arms with higher observed rewards should be selected
often) and exploration (all arms should be explored to learn their average rewards). Equivalently, the performance of a decision rule or algorithm can be measured through its expected regret, defined as the gap between the expected reward achieved by the algorithm and that achieved by an oracle algorithm always selecting the best arm.
MAB problems have found applications in many fields, including sequential clinical trials, communication systems, economics, see e.g. \cite{bubeck_bandit_monograph, cesa2006prediction}.

In this paper, we investigate generic combinatorial MAB problems with linear rewards, as introduced in \cite{cesa2012}. In each round $n\ge 1$, a decision maker selects an arm $M$ from a finite set ${\cal M}\subset\{0,1\}^d$ and receives a reward $M^\top X(n)=\sum_{i=1}^d M_i X_{i}(n)$. The reward vector $X(n)\in \mathbb{R}_+^d$ is unknown. We focus here on the case where all arms consist of the same number $m$ of {\it basic} actions in the sense that $\|M\|_1=m, \;\forall M\in\Mcal$. After selecting an arm $M$ in round $n$, the decision maker receives some feedback. We consider both (i) semi-bandit feedback under which after round $n$, for all $i\in \{1,\ldots,d\}$, the component $X_i(n)$ of the reward vector is revealed if and only if $M_i=1$; (ii) bandit feedback under which only the reward $M^\top X(n)$ is revealed. Based on the feedback received up to round $n-1$, the decision maker selects an arm for the next round $n$, and her objective is to maximize her cumulative reward over a given time horizon consisting of $T$ rounds. The challenge in these problems resides in the very large number of arms, i.e., in its combinatorial structure: the size of $\Mcal$ could well grow as $d^m$. Fortunately, one may hope to exploit the problem structure to speed up the exploration of sub-optimal arms.

We consider two instances of combinatorial bandit problems, depending on how the sequence of reward vectors is generated. We first analyze the case of stochastic rewards, where for all $i\in \{1,\ldots, d\}$, $(X_i(n))_{n\ge 1}$ are i.i.d. with Bernoulli distribution of unknown mean. The reward sequences are also independent across $i$. We then address the problem in the adversarial setting where the sequence of vectors $X(n)$ is arbitrary and selected by an {\it adversary} at the beginning of the experiment. In the stochastic setting, we provide sequential arm selection algorithms whose performance exceeds that of existing algorithms, whereas in the adversarial setting, we devise simple algorithms whose regret have the same scaling as that of state-of-the-art algorithms, but with lower computational complexity.

\section{Contribution and Related Work}\label{sec:sec2}
\subsection{Stochastic combinatorial bandits under semi-bandit feedback}
{\bf Contribution.} (a) We derive an asymptotic (as the time horizon $T$ grows large) regret lower bound satisfied by any algorithm (Theorem \ref{thm:C_theta}).
This lower bound is \emph{problem-specific} and \emph{tight}: there exists an algorithm that attains the bound on all problem instances, although the algorithm might be computationally expensive. To our knowledge, such lower bounds have not been proposed in the case of stochastic combinatorial bandits. The dependency in $m$ and $d$ of the lower bound is unfortunately not explicit. We further provide a simplified lower bound  (Theorem \ref{thm:Ctheta_LB}) and derive its scaling in $(m,d)$ in specific examples.

(b) We propose \algos~ (Efficient Sampling for Combinatorial Bandits), an algorithm whose regret scales at most as $\Ocal(\sqrt{m}d\Delta_{\min}^{-1}\log(T))$ (Theorem \ref{thm:comb-ucb-improved}), where $\Delta_{\min}$ denotes the expected reward difference between the best and the second-best arm. \algos~assigns an index to each arm. The index of given arm can be interpreted as performing likelihood tests with vanishing risk on its average reward. Our indexes are the natural extension of KL-UCB indexes defined for unstructured bandits \cite{garivier2011}. Numerical experiments for some specific combinatorial problems are presented in the supplementary material, and show that \algos~significantly outperforms existing algorithms.

{\bf Related work.} Previous contributions on stochastic combinatorial bandits focused on specific combinatorial structures, e.g. $m$-sets~\cite{anantharam1987asymptotically_iid}, matroids~\cite{kveton2014matroid}, or permutations~\cite{gai2010learning}.  Generic combinatorial problems were investigated in \cite{gai2012combinatorial_TON,chen2013combinatorial_icml,kveton2014tight,wen2015efficient}. The proposed algorithms, \textsc{LLR} and \textsc{CUCB} are variants of the UCB algorithm, and their performance guarantees are presented in Table \ref{table:comparison_stochastic_semibandit}. Our algorithms improve over \textsc{LLR} and \textsc{CUCB} by a multiplicative factor of $\sqrt{m}$.

\begin{table}
\centering
\footnotesize
\begin{tabular}[b]{|c|c|c|c|c|}
\hline
\textbf{Algorithm} &  \textsc{LLR} & \textsc{CUCB}  &
\textsc{CUCB}  & \algos  \\
& \cite{gai2012combinatorial_TON}  &  \cite{chen2013combinatorial_icml} & \cite{kveton2014tight} & (Theorem \ref{thm:comb-ucb-improved})\\ \hline
\textbf{Regret} & $\Ocal\left(\frac{m^3d\Delta_{\max}}{\Delta_{\min}^2}\log(T)\right)$ &
 $\Ocal\left(\frac{m^2d}{\Delta_{\min}}\log(T)\right)$  &
 $\Ocal\left(\frac{md}{\Delta_{\min}}\log(T)\right)$  &
 $\Ocal\left(\frac{\sqrt{m} d}{\Delta_{\min}}\log(T)\right)$ \\ \hline
\end{tabular}
\caption{Regret upper bounds for stochastic combinatorial optimization under semi-bandit feedback.}
\label{table:comparison_stochastic_semibandit}
\normalsize
\end{table}
\subsection{Adversarial combinatorial problems under bandit feedback}
{\bf Contribution.} We present algorithm \textsc{CombEXP}, whose regret is $\Ocal\left(\sqrt{m^3T(d+m^{1/2}\underline{\lambda}^{-1})\log \mu^{-1}_{\min}}\right)$, where \mbox{$\mu_{\min}=\min_{i\in[d]}\frac{1}{m|\Mcal|}\sum_{M\in \Mcal}M_{i}$} and  $\underline{\lambda}$ is the smallest nonzero eigenvalue of the matrix $\EE[MM^\top]$ when $M$ is uniformly distributed over ${\cal M}$ (Theorem \ref{thm:cb2}).
For most problems of interest $m(d\underline{\lambda})^{-1}=\Ocal(1)$ \cite{cesa2012} and $\mu_{\min}^{-1}=\Ocal(\mathrm{poly}(d))$, so that \textsc{CombEXP} has $\Ocal(\sqrt{m^3dT\log(d/m)})$ regret. A known regret lower bound is $\Omega(m\sqrt{dT})$  \cite{audibert2013comb_MoOR}, so the regret gap between \textsc{CombEXP} and this lower bound scales at most as $m^{1/2}$ up to a logarithmic factor.

{\bf Related work.} Adversarial combinatorial bandits have been extensively investigated recently, see \cite{audibert2013comb_MoOR} and references therein. Some papers consider specific instances of these problems, e.g., shortest-path routing \cite{gyorgy2007line}, $m$-sets \cite{kale2010}, and permutations \cite{ailon2014bandit}. For generic combinatorial problems, known regret lower bounds scale as $\Omega\left(\sqrt{mdT}\right)$ and $\Omega\left(m\sqrt{dT}\right)$ (if $d\ge 2m$) in the case of semi-bandit and bandit feedback, respectively \cite{audibert2013comb_MoOR}. In the case of semi-bandit feedback, \cite{audibert2013comb_MoOR} proposes OSMD, an algorithm whose regret upper bound matches the lower bound. \cite{neu2015first} presents an algorithm with $\Ocal(m\sqrt{dL^\star_T\log(d/m)})$ regret where $L^\star_T$ is the total reward of the best arm after $T$ rounds.

For problems with bandit feedback, \cite{cesa2012} proposes \textsc{ComBand} and derives a regret upper bound which depends on the structure of action set $\Mcal$. For most problems of interest, the regret under \textsc{ComBand} is upper-bounded by $\Ocal(\sqrt{m^3dT\log(d/m)})$.
\cite{bubeck2012towards} addresses generic linear optimization with bandit feedback and the proposed algorithm, referred to as \textsc{EXP2 with John's Exploration}, has a regret scaling at most as $\Ocal(\sqrt{m^{3}dT\log(d/m)})$ in the case of combinatorial structure. As we show next, for many combinatorial structures of interest (e.g. $m$-sets, matchings, spanning trees), \textsc{CombEXP} yields the same regret as \textsc{ComBand} and \textsc{EXP2 with John's Exploration}, with lower computational complexity for a large class of problems. Table \ref{table:comparison_adv_bandit} summarises known regret bounds.

\begin{table*}
\centering
\footnotesize
\begin{tabular}[b]{|c|c|}
\hline
\textbf{Algorithm} &  \textbf{Regret} \\ \hline
Lower Bound \cite{audibert2013comb_MoOR} & $\Omega\left(m\sqrt{dT}\right)$, if $d\ge 2m$ \\ \hline
\textsc{ComBand} \cite{cesa2012} & $\Ocal\left(\sqrt{m^3dT\log\frac{d}{m}\left(1+\frac{2m}{d \underline{\lambda}}\right)}\right)$ \\ \hline
\textsc{EXP2 with John's Exploration} \cite{bubeck2012towards} & $\Ocal\left(\sqrt{m^3dT\log \frac{d}{m}}\right)$
\\
\hline
\textsc{CombEXP} (Theorem \ref{thm:cb2}) & $\Ocal\left(\sqrt{m^3dT\left(1+\frac{m^{1/2}}{d \underline{\lambda}}\right) \log\mu_{\min}^{-1}}\right)$ \\ \hline
\end{tabular}
\caption{Regret of various algorithms for adversarial combinatorial bandits with bandit feedback. Note that for most combinatorial classes of interests, $m(d\underline{\lambda})^{-1}=\Ocal(1)$ and $\mu_{\min}^{-1}=\Ocal(\mathrm{poly}(d))$. }
\label{table:comparison_adv_bandit}
\normalsize
\end{table*}

{\bf Example 1: $m$-sets.}
$\Mcal$ is the set of all $d$-dimensional binary vectors with $m$ non-zero coordinates. We have $\mu_{\min}=\frac{m}{d}$ and $\underline{\lambda}=\frac{m(d-m)}{d(d-1)}$ (refer to the supplementary material for details). Hence when $m=o(d)$, the regret upper bound of \textsc{CombEXP} becomes $\Ocal(\sqrt{m^{3}d T\log(d/m)})$, which is the same as that of \textsc{ComBand} and \textsc{EXP2 with John's Exploration}.

{\bf Example 2: matchings.}
The set of arms $\Mcal$ is the set of perfect matchings in $\Kcal_{m,m}$. $d=m^2$ and $|\Mcal|=m!$. We have $\mu_{\min}=\frac{1}{m}$, and $\underline{\lambda}=\frac{1}{m-1}$. Hence the regret upper bound of \textsc{CombEXP} is $\Ocal(\sqrt{m^5 T\log(m)})$, the same as for \textsc{ComBand} and \textsc{EXP2 with John's Exploration}.

{\bf Example 3: spanning trees.}
$\Mcal$ is the set of spanning trees in the complete graph $\Kcal_{N}$. In this case, $d={N\choose 2}$, $m=N-1$, and by Cayley's formula $\Mcal$ has $N^{N-2}$ arms. $\log\mu_{\min}^{-1}\le 2N$ for $N\ge 2$ and $\frac{m}{d\underline{\lambda}}<7$ when $N\ge 6$, The regret upper bound of  \textsc{ComBand} and \textsc{EXP2 with John's Exploration} becomes $\Ocal(\sqrt{N^5T\log(N)})$. As for \textsc{CombEXP}, we get the same regret upper bound $\Ocal(\sqrt{N^{5}T\log(N)})$.

\section{Models and Objectives}
\label{sec:model}
We consider MAB problems where each arm $M$ is a subset of $m$ {\it basic} actions taken from $[d]=\{1,\ldots, d\}$. For $i\in [d]$, $X_i(n)$ denotes the reward of basic action $i$ in round $n$. In the stochastic setting, for each $i$, the sequence of rewards $(X_i(n))_{n\geq 1}$ is i.i.d. with Bernoulli distribution with mean $\theta_i$. Rewards are assumed to be independent across actions. We denote by $\theta=(\theta_1,\dots,\theta_d)^\top\in \Theta=[0,1]^d$ the vector of unknown expected rewards of the various basic actions. In the adversarial setting, the reward vector $X(n)=(X_1(n),\ldots,X_d(n))^\top \in[0,1]^d$ is arbitrary, and the sequence $(X(n), n\ge 1)$ is decided (but unknown) at the beginning of the experiment.

The set of arms $\Mcal$ is an arbitrary subset of $\{0,1\}^d$, such that each of its elements $M$ has $m$ basic actions. Arm $M$ is identified with a binary column vector $(M_1,\dots,M_d)^\top$, and we have $\|M\|_1=m, \;\forall M\in\Mcal$. At the beginning of each round $n$, a policy $\pi$, selects an arm $M^\pi(n)\in\Mcal$ based on the arms chosen in previous rounds and their observed rewards. The reward of arm $M^\pi(n)$ selected in round $n$ is $\sum_{i\in[d]} M^\pi_{i}(n)X_i(n)=M^\pi(n)^\top X(n)$.

We consider both semi-bandit and bandit feedbacks. Under semi-bandit feedback and policy $\pi$, at the end of round $n$, the outcome of basic actions $X_{i}(n)$ for all $i\in M^\pi(n)$ are revealed to the decision maker, whereas under bandit feedback, $M^\pi(n)^\top X(n)$ only can be observed.

Let $\Pi$ be the set of all feasible policies. The objective is to identify a policy in $\Pi$ maximizing the cumulative expected reward over a finite time horizon $T$. The expectation is here taken  with respect to possible randomness in the rewards (in the stochastic setting) and the possible randomization in the policy. Equivalently, we aim at designing a policy that minimizes regret, where the regret of policy $\pi\in\Pi$ is defined by:
$$
R^\pi(T)=\max_{M\in\Mcal} \EE\left[\sum_{n=1}^T M^\top X(n)\right]-\EE\left[\sum_{n=1}^T M^\pi(n)^\top X(n)\right].
$$

Finally, for the stochastic setting, we denote by $\mu_M(\theta)=M^\top  \theta$ the expected reward of arm $M$, and let $M^\star(\theta)\in\Mcal$, or $M^\star$ for short, be any arm with maximum expected reward:
$
M^\star(\theta) \in\arg\max_{M\in\Mcal} \mu_M(\theta).
$
In what follows, to simplify the presentation, we assume that the optimal $M^\star$ is unique. We further define: $\mu^\star(\theta)=M^{\star\top }\theta$, \; $\Delta_{\min}=\min_{M\neq M^\star} \Delta_M$ where $\Delta_M=\mu^\star(\theta)-\mu_M(\theta)$, and $\Delta_{\max}=\max_M(\mu^\star(\theta)-\mu_M(\theta))$.

\section{Stochastic Combinatorial Bandits under Semi-bandit Feedback}

\subsection{Regret Lower Bound}

Given $\theta$, define the set of parameters that cannot be distinguished from $\theta$ when selecting action $M^\star(\theta)$, and for which arm $M^\star(\theta)$ is suboptimal:
\eqs{
B(\theta) = \{ \lambda \in \Theta : M_i^\star(\theta)(\theta_i - \lambda_i) = 0, \ \forall i ,\; \mu^\star(\lambda) > \mu^\star(\theta) \}.
}
We define ${\cal X} = (\RR^+)^{|{\cal M}|}$ and $\klber(u,v)$ the Kullback-Leibler divergence between Bernoulli distributions of respective means $u$ and $v$, i.e., $\klber(u,v)=u\log(u/v)+(1-u)\log( (1-u)/(1-v) )$. Finally, for $(\theta,\lambda) \in \Theta^2$, we define the vector $\klber(\theta,\lambda) = (\klber(\theta_i,\lambda_i) )_{i \in [d]}$.

We derive a regret lower bound valid for any {\it uniformly good} algorithm. An algorithm $\pi$ is uniformly good iff $R^\pi(T) = o(T^\alpha)$ for all $\alpha > 0$ and all parameters $\theta\in \Theta$. The proof of this result relies on a general result on controlled Markov chains \cite{graves1997}.
\begin{theorem}\label{thm:C_theta}
For all $\theta\in \Theta$, for any uniformly good policy $\pi\in \Pi$,\; $
\liminf_{T\to\infty} {R^\pi(T)\over \log(T)} \ge c(\theta),$
where $c(\theta)$ is the optimal value of the optimization problem:
\begin{align}
\label{eq:C_theta_opt}
 \inf_{x \in {\cal X}} \sum_{M \in {\cal M}} x_M (M^\star(\theta) - M)^\top \theta \quad\quad %\label{eq:C_theta_opt_constr}
  \hbox{s.t. }\; \Bigl(\sum_{M \in {\cal M}}  x_M M \Big)^\top \klber(\theta,\lambda) \ge 1 \;,\; \forall \lambda \in B(\theta).
\end{align}
\end{theorem}

Observe first that optimization problem (\ref{eq:C_theta_opt}) is a semi-infinite linear program which can be solved for any fixed $\theta$, but its optimal value is difficult to compute explicitly. Determining how $c(\theta)$ scales as a function of the problem dimensions $d$ and $m$ is not obvious. Also note that (\ref{eq:C_theta_opt}) has the following interpretation: assume that (\ref{eq:C_theta_opt}) has a unique solution $x^\star$. Then any uniformly good algorithm must select action $M$ at least $x^\star_M\log(T)$ times over the $T$ first rounds. From \cite{graves1997}, we know that there exists an algorithm which is asymptotically optimal, so that its regret matches the lower bound of Theorem~\ref{thm:C_theta}. However this algorithm suffers from two problems: it is computationally infeasible for large problems since it involves solving (\ref{eq:C_theta_opt}) $T$ times, furthermore the algorithm has no finite time performance guarantees, and numerical experiments suggests that its finite time performance on typical problems is rather poor. Further remark that if ${\cal M}$ is the set of singletons (classical bandit), Theorem~\ref{thm:C_theta} reduces to the Lai-Robbins bound \cite{lai1985} and if ${\cal M}$ is the set of $m$-sets (bandit with multiple plays), Theorem~\ref{thm:C_theta} reduces to the lower bound derived in \cite{anantharam1987asymptotically_iid}. Finally, Theorem~\ref{thm:C_theta} can be generalized in a straightforward manner for when rewards belong to a one-parameter exponential family of distributions (e.g., Gaussian, Exponential, Gamma etc.) by replacing $\klber$ by the appropriate divergence measure.

\paragraph{A Simplified Lower Bound}
We now study how the regret $c(\theta)$ scales as a function of the problem dimensions $d$ and $m$. To this aim, we present a simplified regret lower bound. Given $\theta$, we say that a set $\Hcal \subset \Mcal\setminus M^\star$ has property $P(\theta)$ iff, for all $(M,M') \in {\cal H}^2$, $M \neq M'$ we have $M_i M'_i (1 - M^\star_i(\theta)) = 0$ for all $i$. We may now state Theorem~\ref{thm:Ctheta_LB}.

\begin{theorem}
\label{thm:Ctheta_LB}
Let $\Hcal$ be a maximal (inclusion-wise) subset of $\Mcal$ with property $P(\theta)$. Define $\beta(\theta)=\min_{M\neq M^\star}\frac{\Delta_M}{|M\setminus M^\star|}$. Then:
\begin{align*}
c(\theta)\ge \sum_{M\in \Hcal}\frac{\beta(\theta)}{\max_{i\in M\setminus M^\star} \klber\left(\theta_i,\frac{1}{|M\setminus M^\star|}\sum_{j\in M^\star\setminus M}\theta_j\right)}.
\end{align*}
\end{theorem}

\begin{corollary}\label{cor:Ctheta_LB}
Let $\theta\in [a,1]^d$ for some constant $a>0$ and $\Mcal$ be such that each arm $M\in \Mcal, M\neq M^\star$ has at most $k$ suboptimal basic actions. Then $c(\theta)=\Omega(|\Hcal|/k)$.
\end{corollary}
Theorem~\ref{thm:Ctheta_LB} provides an explicit regret lower bound. Corollary \ref{cor:Ctheta_LB} states that $c(\theta)$ scales at least with the size of $\Hcal$. For most combinatorial sets, $|\Hcal|$ is proportional to $d-m$ (see supplementary material for some examples), which implies that in these cases, one cannot obtain a regret smaller than $\Ocal( (d-m) \Delta_{\min}^{-1}\log(T))$. This result is intuitive since $d-m$ is the number of parameters not observed when selecting the optimal arm. The algorithms proposed below have a regret of $\Ocal(d \sqrt{m}\Delta_{\min}^{-1}\log(T))$, which is acceptable since typically, $\sqrt{m}$ is much smaller than $d$.
\subsection{Algorithms}
\label{sec:algos}
Next we present \algos, an algorithm for stochastic combinatorial bandits that relies on arm indexes as in UCB1 \cite{auer2002} and KL-UCB \cite{garivier2011}. We derive finite-time regret upper bounds for \algos~that hold even if we assume that $\|M\|_1\le m,\;\forall M\in\Mcal$, instead of $\|M\|_1 =m$, so that arms may have different numbers of basic actions.

\subsubsection{Indexes}

\algos~relies on arm indexes. In general, an index of arm $M$ in round $n$, say $b_M(n)$, should be defined so that $b_M(n) \geq M^\top \theta$ with high probability. Then as for UCB1 and KL-UCB, applying the principle of optimism against uncertainty, a natural way  to devise algorithms based on indexes is to select in each round the arm with the highest index. Under a given algorithm, at time $n$, we define $t_i(n) = \sum_{s=1}^n M_i(s)$ the number of times basic action $i$ has been sampled. The empirical mean reward of action $i$ is then defined as $\hat\theta_i(n) = (1/t_i(n)) \sum_{s=1}^n X_i(s) M_i(s)$ if $t_i(n) > 0$  and $\hat\theta_i(n) = 0$ otherwise. We define the corresponding vectors $t(n) = (t_i(n))_{i \in [d]}$ and $\hat\theta(n) = ( \hat\theta_i(n) )_{i \in [d]}$.

The indexes we propose are functions of the round $n$ and of $\hat\theta(n)$. Our first index for arm $M$, referred to as $b_M(n,\hat\theta(n))$ or $b_M(n)$ for short, is an extension of KL-UCB index. Let $f(n) = \log(n) + 4 m \log(\log(n))$. $b_M(n,\hat\theta(n))$ is the optimal value of the following optimization problem:
\al{\label{eq:optim_problem}
\max_{ q \in \Theta }  M^\top q \quad\quad \text{s.t. } \;\; (Mt(n))^\top \klber(\hat\theta(n),q) \leq f(n),
}
where we use the convention that for $v,u \in \RR^d$, $vu = (v_i u_i)_{i \in [d]}$.
%Index $b_M(n)$ can be interpreted as performing likelihood tests with vanishing risk (more precisely $\Ocal(n^ {-1})$) on the average reward of arm $M$.
As we show later, $b_M(n)$ may be computed efficiently using a line search procedure similar to that used to determine KL-UCB index.

Our second index $c_M(n,\hat\theta(n))$ or $c_M(n)$ for short is a generalization of the UCB1 and UCB-tuned indexes:
\eqs{
c_M(n) = M^\top \hat\theta(n) + \sqrt{\frac{f(n)}{2} \left( \sum_{i=1}^{d} \frac{M_i}{t_i(n)} \right)}
}
Note that, in the classical bandit problems with independent arms, i.e., when $m = 1$, $b_M(n)$ reduces to the KL-UCB index (which yields an asymptotically optimal algorithm) and $c_M(n)$ reduces to the UCB-tuned index. The next theorem provides generic properties of our indexes. An important consequence of these properties is that the expected number of times where $b_{M^\star}(n,\hat\theta(n))$ or $c_{M^\star}(n,\hat\theta(n))$ underestimate $\mu^\star(\theta)$ is \emph{finite}, as stated in the corollary below.
\begin{theorem}\label{th:indexes}
(i) For all $n\ge 1$, $M\in {\cal M}$ and $\tau\in [0,1]^d$, we have $b_M(n,\tau) \leq c_M(n,\tau)$.\\
(ii) There exists $C_m > 0$  depending on $m$ only such that, for all $M\in {\cal M}$ and $n\ge 2$:
$$
\PP[b_M(n,\hat\theta(n)) \leq M^\top \theta] \leq C_m n^{-1} (\log(n))^{-2}.
$$
\end{theorem}
\begin{corollary} $
\sum_{n \geq 1} \PP[b_{M^\star}(n,\hat\theta(n)) \leq \mu^\star ]  \leq1+ C_m\sum_{n \geq 2}  n^{-1} (\log(n))^{-2} < \infty.
$
\end{corollary}

Statement (i) in the above theorem is obtained combining Pinsker and Cauchy-Schwarz inequalities. The proof of statement (ii) is based on a concentration inequality on sums of empirical KL divergences proven in \cite{magureanu2014lipschitz}. It enables to control the fluctuations of multivariate empirical distributions for exponential families. It should also be observed that indexes $b_M(n)$ and $c_M(n)$ can be extended in a straightforward manner to the case of continuous linear bandit problems, where the set of arms is the unit sphere and one wants to maximize the dot product between the arm and an unknown vector. $b_M(n)$ can also be extended to the case where reward distributions are not Bernoulli but lie in an exponential family (e.g. Gaussian, Exponential, Gamma, etc.), replacing $\klber$ by a suitably chosen divergence measure. A close look at $c_M(n)$ reveals that the indexes proposed in \cite{chen2013combinatorial_icml}, \cite{kveton2014tight}, and \cite{gai2012combinatorial_TON} are too conservative to be optimal in our setting: there the ``confidence bonus''  $\sum_{i=1}^{d} \frac{M_i}{t_i(n)}$ was replaced by (at least) $m \sum_{i=1}^{d} \frac{M_i}{t_i(n)}$.
Note that \cite{chen2013combinatorial_icml}, \cite{kveton2014tight} assume that the various basic actions are arbitrarily correlated, while we assume independence among basic actions. When independence does not hold, \cite{kveton2014tight} provides a problem instance where the regret is at least $\Ocal(\frac{md}{\Delta_{\min}}\log(T))$. This does not contradict our regret upper bound (scaling as $\Ocal(\frac{d\sqrt{m}}{\Delta_{\min}}\log(T))$), since we have added the independence assumption.

\subsubsection{Index computation}

While the index $c_M(n)$ is explicit, $b_M(n)$ is defined as the solution to an optimization problem. We show that it may be computed by a simple line search. For $\lambda \geq 0$, $w \in [0,1]$ and $v \in \NN$, define:
\eqs{
g(\lambda,w,v) = \left( 1 - \lambda v + \sqrt{ (1 - \lambda v)^2 + 4 w v \lambda}\right)/2.
}
Fix $n$, $M$, $\hat\theta(n)$ and $t(n)$. Define $I= \{ i: M_i = 1, \hat\theta_i(n) \neq 1\}$, and for $\lambda > 0$, define:
\eqs{
F(\lambda) = \sum_{i \in I} t_i(n) \klber( \hat\theta_i(n), g(\lambda,\hat\theta_i(n),t_i(n))).
}
\begin{theorem}\label{th:index_computation}
If $I = \emptyset$, $b_M(n) = ||M||_1$. Otherwise: 	
    (i) $\lambda \mapsto F(\lambda)$ is strictly increasing, and $F(\RR^+) = \RR^+$.
	(ii) Define $\lambda^\star$ as the unique solution to $F(\lambda) = f(n)$. Then $b_M(n) = ||M||_1 - |I| + \sum_{i \in I} g(\lambda^\star,\hat\theta_i(n),t_i(n))$.
\end{theorem}
Theorem~\ref{th:index_computation} shows that $b_M(n)$ can be computed using a line search procedure such as bisection, as this computation amounts to solving the nonlinear equation $F(\lambda) = f(n)$, where $F$ is strictly increasing. The proof of Theorem~\ref{th:index_computation} follows from KKT conditions and the convexity of  KL divergence.

\subsubsection{The \algos~Algorithm}
\label{sec:ucb_bestindex}

The pseudo-code of \algos~is presented in Algorithm \ref{alg:algo_stochastic}. We consider two variants of the algorithm based on the choice of the index $\xi_M(n)$: \algos-1 when $\xi_M(n) = b_M(n)$ and \algos-2 if $\xi_M(n) = c_M(n)$. In practice, \algos-1~outperforms \algos-2. Introducing \algos-2~is however instrumental in the regret analysis of \algos-1~(in view of Theorem \ref{th:indexes} (i)).
The following theorem provides a finite time analysis of our \algos~algorithms. The proof of this theorem borrows some ideas from the proof of \cite[Theorem~3]{kveton2014tight}.

\begin{algorithm}[tb]
\small
   \caption{\algos}
   \label{alg:algo_stochastic}
\begin{algorithmic}
   %\STATE {\bf Initialization:} For $n=1,\ldots, A$, select actions in ${\cal A}$, observe the rewards, and update $\zeta(n)$.
   %\vspace{1mm}
   \FOR{$n\geq 1$}
   \STATE Select arm $M(n)\in \arg\max_{M\in {\cal M}} \xi_M(n)$. %\vspace{1mm}
   \STATE Observe the rewards, and update $t_i(n)$ and $\hat\theta_i(n), \forall i\in M(n)$. %\vspace{1mm}
   \ENDFOR
\end{algorithmic}
\normalsize
\end{algorithm}

\begin{theorem}\label{thm:comb-ucb-improved}
The regret under algorithms $\pi \in \{\algos\textsc{-1} , \algos\textsc{-2}\}$ satisfies for all $T\ge 1$:
\eqs{
R^{\pi}(T) \leq  16d\sqrt{m}\Delta_{\min}^{-1}f(T) + 4dm^3\Delta_{\min}^{-2} + C'_m,
}
where $C'_m \geq 0$  does not depend on $\theta$, $d$ and $T$. As a consequence $R^{\pi}(T) = \Ocal(d\sqrt{m} \Delta_{\min}^{-1} \log(T))$ when $T \to \infty$.
\end{theorem}

\algos~with time horizon $T$ has a complexity of $\Ocal(|\Mcal| T)$ as neither $b_M$ nor $c_M$ can be written as $M^\top y$ for some vector $y\in\mathbb R^d$. Assuming that the offline (static) combinatorial problem is solvable in $\Ocal(\OPT(\Mcal))$ time, the complexity of CUCB algorithm in \cite{chen2013combinatorial_icml} and \cite{kveton2014tight} after $T$ rounds is $\Ocal(\OPT(\Mcal) T)$. Thus, if the offline problem is efficiently implementable, i.e., $\OPT(\Mcal)=\Ocal(\mathrm{poly}(d))$, CUCB is efficient, whereas \algos~is not since $|\Mcal|$ may have exponentially many elements. In \textsection 2.5 of the supplement, we provide an extension of \algos~called \textsc{Epoch-}\algos, that attains almost the same regret as \algos~while enjoying much better computational complexity.

\section{Adversarial Combinatorial Bandits under Bandit Feedback}
\label{sec:adv_comb_bandit}
We now consider adversarial combinatorial bandits with bandit feedback.
We start with the following observation:
\begin{align*}
\max_{M\in \Mcal} M^\top X = \max_{\mu \in Co(\Mcal)} \mu^\top X,
\end{align*}
with $Co(\Mcal)$ the convex hull of $\Mcal$. We embed $\mathcal{M}$ in the $d$-dimensional simplex by dividing its elements by $m$.
Let $\Pcal$ be this scaled version of $Co(\Mcal)$.

Inspired by OSMD \cite{audibert2013comb_MoOR,bubeck2012towards}, we propose the \textsc{CombEXP} algorithm, where the KL divergence is the Bregman divergence used to project onto $\Pcal$. Projection using the KL divergence is addressed in~\cite{csishi04}. We denote the KL divergence between distributions $q$ and $p$ in $\Pcal$ by
\mbox{$
\; \kl(p,q)=\sum_{i\in[d]} p(i)\log \frac{p(i)}{q(i)}.
$}
The projection of distribution $q$ onto a closed convex set $\Xi$ of distributions is \mbox{$p^\star=\arg\min_{p\in \Xi}\kl(p,q).$}

Let $\underline{\lambda}$ be the smallest nonzero eigenvalue of $\EE[MM^\top]$, where $M$ is uniformly distributed over ${\cal M}$. We define the exploration-inducing distribution $\mu^0\in\Pcal$:
$
\mu^0_{i}=\frac{1}{m|\Mcal|}\sum_{M\in \Mcal}M_{i},\quad \forall i\in[d],
$
and let $\mu_{\min}=\min_{i} m\mu^0_{i}.$ $\mu^0$ is the distribution over basic actions $[d]$ induced by the uniform distribution over $\Mcal$. The pseudo-code for \textsc{CombEXP} is shown in Algorithm~\ref{alg:CombEXP}. The KL projection in \textsc{CombEXP} ensures that $mq_{n-1}\in Co(\Mcal)$. There exists $\lambda$, a distribution over ${\cal M}$ such that $mq_{n-1}=\sum_{M} \lambda(M) M$. This guarantees that the system of linear equations in the \emph{decomposition step} is consistent.
We propose to perform the \emph{projection step} (the KL projection of $\tilde{q}$ onto $\Pcal$) using interior-point methods \cite{boyd2004convex}. We provide a simpler method in \textsection 3.4 of the supplement. The \emph{decomposition step} can be efficiently implemented using the algorithm of \cite{sherali1987constructive}. The following theorem provides a regret upper bound for \textsc{CombEXP}.

\begin{algorithm}[tb]
\small
   \caption{\textsc{CombEXP}}
   \label{alg:CombEXP}
\begin{algorithmic}
   \STATE {\bf Initialization:} Set $q_0=\mu^0$, $\gamma=\frac{\sqrt{m\log \mu_{\min}^{-1}}} {\sqrt{m\log \mu_{\min}^{-1}}+\sqrt{C(Cm^2d+m)T}}$ and $\eta=\gamma C$, with $C=\frac{\underline{\lambda}}{m^{3/2}}$. %\vspace{2mm}
   \FOR{$ n\ge  1$}
   \STATE \emph{Mixing:} Let $q'_{n-1}=(1-\gamma)q_{n-1}+\gamma\mu^0$. \vspace{.5mm}
   \STATE \emph{Decomposition:} Select a distribution $p_{n-1}$ over $\Mcal$ such that $\sum_{M} p_{n-1}(M) M=mq'_{n-1}$. \vspace{.5mm}
   \STATE \emph{Sampling:} Select a random arm $M(n)$ with distribution $p_{n-1}$ and incur a reward $Y_n=\sum_{i}X_i(n)M_{i}(n)$.  %\vspace{.5mm}
   \STATE \emph{Estimation:} Let $\Sigma_{n-1}=\EE\left[ MM^{\top}\right]$, where $M$ has law $p_{n-1}$. Set $\tilde{X}(n) = Y_n\Sigma_{n-1}^{+}M(n)$, where  $\Sigma_{n-1}^{+}$ is the pseudo-inverse of $\Sigma_{n-1}$. \vspace{.5mm}
   \STATE \emph{Update:} Set $\tilde{q}_n(i) \propto q_{n-1}(i) \exp(\eta \tilde{X}_{i}(n)),\; \forall i\in[d]$. \vspace{.5mm}
   \STATE \emph{Projection:} Set $q_n$ to be the projection of $\tilde{q}_n$ onto the set $\Pcal$ using the KL divergence. %\vspace{1.5mm}
   \ENDFOR
\end{algorithmic}
\normalsize
\end{algorithm}

\begin{theorem}
\label{thm:cb2}
For all $T\ge 1$\emph{:}
%\begin{align*}
$
R^{\textsc{CombEXP}}(T) \le 2\sqrt{m^3T\left(d+\frac{m^{1/2}}{\underline{\lambda}}\right) \log \mu_{\min}^{-1}}
                    +\frac{m^{5/2}}{\underline{\lambda}}\log\mu_{\min}^{-1}.$
%\end{align*}
\end{theorem}

For most classes of $\Mcal$, we have $\mu_{\min}^{-1}=\Ocal(\mathrm{poly}(d))$ and \mbox{$m(d\underline{\lambda})^{-1}=\Ocal(1)$} \cite{cesa2012}. For these classes, \textsc{CombEXP} has a regret of $\Ocal(\sqrt{m^{3}dT\log(d/m)})$, which is a factor $\sqrt{m \log (d/m)}$ off the lower bound (see Table \ref{table:comparison_adv_bandit}).

It might not be possible to compute the projection step exactly, and this step can be solved up to accuracy $\epsilon_n$ in round $n$. Namely we find $q_n$ such that \mbox{$\kl(q_n,\tilde q_n) - \min_{p\in \Xi}\kl(p,\tilde q_n)  \leq \epsilon_n$}. Proposition~\ref{prop:epsilon} shows that for $\epsilon_n = \Ocal(n^{-2}\log^{-3}(n))$, the approximate projection gives the same regret as when the projection is computed exactly. Theorem~\ref{thm:cb2-complexity} gives the computational complexity of \textsc{CombEXP} with approximate projection. When $Co(\Mcal)$ is described by polynomially (in $d$) many linear equalities/inequalities, \textsc{CombEXP} is efficiently implementable and its running time scales (almost) linearly in $T$. Proposition \ref{prop:epsilon} and Theorem \ref{thm:cb2-complexity} easily extend to other OSMD-type algorithms and thus might be of independent interest.

%Let $\underline{q}_n=\min_{i} q_n(i)$.
\begin{proposition}\label{prop:epsilon}
 If the projection step of \textsc{CombEXP} is solved up to accuracy \mbox{$\epsilon_n=\Ocal(n^{-2}\log^{-3}(n))$}, we have:
$$
R^{\textsc{CombEXP}}(T) \le 2\sqrt{2m^3T\left(d+\frac{m^{1/2}}{\underline{\lambda}}\right) \log \mu_{\min}^{-1}}+\frac{2m^{5/2}}{\underline{\lambda}}\log\mu_{\min}^{-1}.
$$
\end{proposition}

\begin{theorem}
\label{thm:cb2-complexity}
Assume that $Co({\cal M})$ is defined by $c$ linear equalities and $s$ linear inequalities. If the projection step is solved up to accuracy $\epsilon_n=\Ocal(n^{-2}\log^{-3}(n))$, then \textsc{CombEXP} has time complexity $\Ocal(T[ \sqrt{s}(c+d)^3\log(T)+d^4])$.
\end{theorem}

The time complexity of \textsc{CombEXP} can be reduced by exploiting the structure of ${\cal M}$ (See \cite[page~545]{boyd2004convex}). In particular, if inequality constraints describing $Co(\Mcal)$ are box constraints, the time complexity of \textsc{CombEXP} is $\Ocal(T[ c^2\sqrt{s}(c+d)\log(T)+d^4])$.

The computational complexity of \textsc{CombEXP} is determined by the structure of $Co(\Mcal)$ and \textsc{CombEXP} has $\Ocal(T\log(T))$ time complexity due to the efficiency of interior-point methods. In contrast, the computational complexity of \textsc{ComBand} depends on the complexity of sampling from $\Mcal$. \textsc{ComBand} may have a time complexity that is super-linear in $T$ (see \cite[page~217]{ailon2014bandit}).  For instance, consider the matching problem described in Section~\ref{sec:sec2}. We have $c=2m$ equality constraints and $s=m^2$ box constraints, so that the time complexity of \textsc{CombEXP} is: $\Ocal(m^5T\log(T))$. It is noted that using \cite[Algorithm~1]{helmbold2009}, the cost of decomposition in this case is $\Ocal(m^4)$. On the other hand, \textsc{CombBand} has a time complexity of $\Ocal(m^{10}F(T))$, with $F$ a super-linear function, as it requires to approximate a permanent, requiring $\Ocal(m^{10})$ operations per round. Thus, \textsc{CombEXP} has much lower complexity than \textsc{ComBand} and achieves the same regret.

\section{Conclusion}
\label{sec:conclusion}
We have investigated stochastic and adversarial combinatorial bandits. For stochastic combinatorial bandits with semi-bandit feedback, we have provided a tight, problem-dependent regret lower bound that, in most cases, scales at least as $\Ocal((d-m)\Delta_{\min}^{-1}\log(T))$. We proposed~\algos, an algorithm with $\Ocal(d\sqrt{m}\Delta_{\min}^{-1}\log(T))$ regret. We plan to reduce the gap between this regret guarantee and the regret lower bound, as well as investigate the performance of \textsc{Epoch-}\algos. For adversarial combinatorial bandits with bandit feedback, we proposed the \textsc{CombEXP} algorithm. There is a gap between the regret of \textsc{CombEXP} and the known regret lower bound in this setting, and we plan to reduce it as much as possible.

\subsubsection*{Acknowledgments}
A. Proutiere's research is supported by the ERC FSA grant, and the SSF ICT-Psi project.
\newpage

%\subsubsection*{References}

\small
\bibliography{bandit}
\bibliographystyle{unsrt}

\newpage

%\maketitle

\appendix

\begin{center}
\textbf{\LARGE Supplementary Materials and Proofs}
\end{center}
\normalsize

\section{Stochastic Combinatorial Bandits: Regret Lower Bounds}
\label{sec:supp_LB}

\subsection{Proof of Theorem 1} %Theorem \ref{thm:C_theta}

To derive regret lower bounds, we apply the techniques used by Graves and Lai \cite{graves1997} to investigate efficient adaptive decision rules in controlled Markov chains. First we give an overview of their general framework.

Consider a controlled Markov chain $(X_n)_{n\ge 0}$ on a finite state space $\mathcal S$ with a control set $U$. The transition probabilities given control $u\in U$ are parameterized by $\theta$ taking values in a compact metric space $\Theta$: the probability to move from state $x$ to state $y$ given the control $u$ and the parameter $\theta$ is $p(x,y;u,\theta)$. The parameter $\theta$ is not known. The decision maker is provided with a finite set of stationary control laws $G=\{g_1,\ldots,g_K\}$, where each control law $g_j$ is a mapping from $\mathcal S$ to $U$: when control law $g_j$ is applied in state $x$, the applied control is $u=g_j(x)$. It is assumed that if the decision maker always selects the same control law $g$, the Markov chain is then irreducible with stationary distribution $\pi_\theta^g$. Now the reward obtained when applying control $u$ in state $x$ is denoted by $r(x,u)$, so that the expected reward achieved under control law $g$ is: $\mu_\theta(g)=\sum_xr(x,g(x))\pi_\theta^g(x)$. There is an optimal control law given $\theta$ whose expected reward is denoted by $\mu_\theta^{\star}=\max_{g\in G} \mu_\theta(g)$. Now the objective of the decision maker is to sequentially select control laws so as to maximize the expected reward up to a given time horizon $T$. As for MAB problems, the performance of a decision scheme can be quantified through the notion of regret which compares the expected reward to that obtained by always applying the optimal control law.\newline

\bp
The parameter $\theta$ takes values in $[0,1]^d$. The Markov chain has values in $\mathcal S=\{0,1\}^d$. The set of controls corresponds to the set of feasible actions $\mathcal M$, and the set of control laws is also $\Mcal$. These laws are constant, in the sense that the control applied by control law $M\in \Mcal$ does not depend on the state of the Markov chain, and corresponds to selecting action $M$. The transition probabilities are given as follows: for all $x,y\in \mathcal S$,
$$
p(x,y;M,\theta)=p(y;M,\theta)=\prod_{i\in [d]}p_i(y_i;M,\theta),
$$
where for all $i\in [d]$, if $M_i=0$, $p_i(0;M,\theta)=1$, and if $M_i=1$, $p_i(y_i;M,\theta)=\theta_{i}^{y_i}(1-\theta_{i})^{1-y_i}$. Finally, the reward $r(y,M)$ is defined by $r(y,M)=M^\top y$. Note that the state space of the Markov chain is here finite, and so, we do not need to impose any cost associated with switching control laws (see the discussion on page 718 in \cite{graves1997}).

We can now apply Theorem 1 in \cite{graves1997}. Note that the KL number under action $M$ is
$$
\klber^M(\theta,\lambda) =\sum_{i\in [d]} M_i\klber(\theta_{i},\lambda_{i}).
$$
From \cite[Theorem~1]{graves1997}, we conclude that for any uniformly good rule $\pi$,
$$
\liminf_{T\to\infty} {R^\pi(T)\over \log(T)} \ge c(\theta),
$$
where $c(\theta)$ is the optimal value of the following optimization problem:
\begin{align}
\label{eq:C_theta_opt}
& \inf_{x_M\ge 0, M\in \Mcal} \sum_{M\neq M^\star} x_M(\mu^\star - \mu_M(\theta)),\\
\label{eq:C_theta_opt_constr}
& \hbox{s.t. }\inf_{\lambda\in B(\theta)} \sum_{Q\neq M^\star}x_Q\klber^Q(\theta,\lambda) \ge 1.
\end{align}
The result is obtained by observing that $B(\theta)=\bigcup_{M\neq M^\star}B_M(\theta)$, where
\begin{align*}
B_M(\theta)=\{ \lambda\in \Theta : M^\star_i(\theta)(\theta_i-\lambda_i)=0, \forall i, \; \mu^\star(\theta) < \mu_M(\lambda)\}.
\end{align*}

\ep

\subsection{Proof of Theorem 2} %Theorem \ref{thm:Ctheta_LB}

The proof proceeds in three steps. In the subsequent analysis, given the optimization problem \textsf{P}, we use $\textrm{val}(\textsf{P})$ to denote its optimal value.

\paragraph{\underline{Step 1}.}
In this step, first we introduce an equivalent formulation for problem (\ref{eq:C_theta_opt}) above  by simplifying its constraints. We show that constraint (\ref{eq:C_theta_opt_constr}) is equivalent to:
\begin{align*}
\inf_{\lambda\in B_M(\theta)}\sum_{i\in M\setminus M^\star}\klber(\theta_i,\lambda_i)\sum_{Q\in \Mcal} Q_ix_Q\ge 1,\;\; \forall M\neq M^\star.\nonumber
\end{align*}
Observe that:
\begin{align*}
\sum_{Q\neq M^\star} x_Q\klber^Q(\theta,\lambda)=\sum_{Q\neq M^\star} x_Q\sum_{i\in[d]} Q_{i}\klber(\theta_{i},\lambda_{i})
=\sum_{i\in[d]}\klber(\theta_i,\lambda_i)\sum_{Q\neq M^\star} Q_i x_Q.
\end{align*}
Fix $M\neq M^\star$. In view of the definition of $B_M(\theta)$, we can find $\lambda\in B_M(\theta)$ such that $\lambda_i=\theta_i, \forall i\in ([d]\setminus M)\cup M^\star$. Thus, for the r.h.s. of the $M$-th constraint in (\ref{eq:C_theta_opt_constr}), we get:
\begin{align}
\inf_{\lambda\in B_M(\theta)} \sum_{Q\neq M^\star}x_Q\klber^Q(\theta,\lambda) &=
\inf_{\lambda\in B_M(\theta)} \sum_{i\in[d]}\klber(\theta_i,\lambda_i)\sum_{Q\neq M^\star} Q_i x_Q\nonumber\\
&= \inf_{\lambda\in B_M(\theta)}\sum_{i\in M\setminus M^\star}\klber(\theta_i,\lambda_i)\sum_{Q} Q_ix_Q,\nonumber
\end{align}
and therefore problem (\ref{eq:C_theta_opt}) can be equivalently written as:
\begin{align}
\label{eq:C_theta_opt_alter}
c(\theta)=& \inf_{x_M\ge 0, M\in {\cal M}} \sum_{M\neq M^\star} x_M(\mu^\star - \mu_M(\theta)),\\
&  \hbox{s.t. }\inf_{\lambda\in B_M(\theta)}\sum_{i\in M\setminus M^\star}\klber(\theta_i,\lambda_i)\sum_{Q} Q_ix_Q \ge 1,\;\; \forall M\neq M^\star.
\end{align}

Next, we formulate an LP whose value gives a lower bound for $c(\theta)$. % Let $x^\star$ be the optimal solution of problem (\ref{eq:C_theta_opt_alter}).
%Then, for any $M\neq M^\star$ define
%$$
%\lambda^\star(M)\in\arg\inf_{\lambda\in B_M(\theta)}\; \sum_{i\in M\setminus M^\star} \klber(\theta_i,\lambda_i)\sum_{Q} Q_ix^\star_Q,
%$$
%where the uniqueness of the minimizer is due to strict convexity of $\klber(\cdot,\cdot)$ in its second argument.
Define $\hat\lambda(M)=(\hat\lambda_i(M), i\in[d])$ with
\[ \hat\lambda_{i}(M) = \left\{
  \begin{array}{l l}
    \frac{1}{|M\setminus M^\star|}\sum_{j\in M^\star\setminus M}\theta_j & \quad \hbox{if }i\in M\setminus M^\star,\vspace{1.5mm}\\
    \theta_i & \quad \textrm{otherwise.}
  \end{array} \right.\]
Clearly $\hat\lambda(M)\in B_M(\theta)$, and therefore:
\begin{align*}
\inf_{\lambda\in B_M(\theta)}\; \sum_{i\in M\setminus M^\star} \klber(\theta_i,\lambda_i)\sum_{Q} Q_ix_Q&\le \sum_{i\in M\setminus M^\star} \klber(\theta_i,\hat\lambda_i(M))\sum_{Q} Q_ix_Q,
\end{align*}

Then, we can write:
\begin{align}
\label{eq:Ctheta_opt_alter2}
c(\theta)\ge& \inf_{x\ge 0} \sum_{M\neq M^\star} \Delta_Mx_M\\
& \hbox{s.t. } \sum_{i\in M\setminus M^\star} \klber(\theta_i,\hat\lambda_i(M))
\sum_{Q} Q_ix_Q \ge 1,\;\;\; \forall M\neq M^\star.
\end{align}

%Fix $M\neq M^\star$ and introduce \mbox{$\hat\lambda(M)=(\hat\lambda_{l}(M),i\in[N_1]\times [N_2])$} with
%\[ \hat\lambda_{l}(M) = \left\{
%  \begin{array}{l l}
%    \frac{1}{|M\setminus M^\star|}\sum_{k\in M^\star\setminus M} \theta_{k}	& \quad i\in M\setminus M^\star\\
%		\theta_i		&		\textrm{otherwise}
%  \end{array} \right.\]
%In view of the definition of $B_M(\theta)$, it can be easily seen that $\hat\lambda(M)\in B_M(\theta)$, and thus:
%$$\sum_{i\in M\setminus M^\star} \klber(\theta_i,\hat\lambda_i(M))\sum_{Q} Q_ix^\star_Q\ge  \sum_{i\in M\setminus M^\star} \klber(\theta_i,\lambda^\star_l(M))\sum_Q Q_ix^\star_Q.
%$$
For any $M\neq M^\star$ introduce: $g_M=\max_{i\in M\setminus M^\star} \klber(\theta_i,\hat\lambda_i(M))$. Now we form \textsf{P1} as follows:
\begin{align}
\textsf{P1:}\quad& \inf_{x\ge 0} \;\sum_{M\neq M^\star} \Delta_M x_M\\
& \hbox{s.t. } \sum_{i \in M\setminus M^\star} \sum_{Q} Q_ix_Q\geq \frac{1}{g_M},\quad \forall M\neq M^\star.
\end{align}
Observe that $c(\theta)\ge \textrm{val}(\textsf{P1})$ since the feasible set of problem (\ref{eq:Ctheta_opt_alter2}) is contained in that of \textsf{P1}.

\paragraph{\underline{Step 2.}}
In this step, we formulate an LP to give a lower bound for val(\textsf{P1}).
To this end, for any suboptimal basic action $i\in [d]$, we define $z_i=\sum_{M} M_ix_M$. Further, we let  $z=[z_i,i\in[d]]$.
Next, we represent the objective of \textsf{P1} in terms of $z$, and give a lower bound for it as follows:

\begin{align*}
\sum_{M\neq M^\star} \Delta_M x_M&= \sum_{M\neq M^\star} x_M\sum_{i\in M\setminus M^\star} \frac{\Delta_M}{|M\setminus M^\star|} \\
						&=\sum_{M\neq M^\star} x_M\sum_{i\in [d]\setminus M^\star} \frac{\Delta_M}{|M\setminus M^\star|}M_i \\
			&\ge\min_{M\neq M^\star} \frac{\Delta_M}{|M\setminus M^\star|}\cdot\sum_{i\in [d]\setminus M^\star} \sum_{M'\neq M^\star} M'_ix_{M'}  \\
			&=\min_{M\neq M^\star} \frac{\Delta_M}{|M\setminus M^\star|}\cdot\sum_{i\in [d]\setminus M^\star} z_{i}\\
            &= \beta(\theta) \sum_{i\in [d]\setminus M^\star} z_{i}.
\end{align*}
Then, defining
\begin{align*}
\textsf{P2:}&\quad  \inf_{z\ge 0} \beta(\theta)\sum_{i\in [d]\setminus M^\star} z_i\\
& \hbox{s.t. } \sum_{i\in M\setminus M^\star} z_{i}\geq \frac{1}{g_M},\;\; \forall M\neq M^\star,
\end{align*}
yields:
$\textrm{val}(\textsf{P1})\ge \textrm{val}(\textsf{P2}).$

\paragraph{\underline{Step 3.}}
%Now, Let $\Hcal$ be a maximal subset (with respect to inclusion) of $\Mcal$ that satisfies the following property:
%$$
%\forall M,M'\in \Hcal, \; M\neq M'\;\Longrightarrow\; (M\setminus M^\star)\cap (M'\setminus M^\star)=\emptyset.
%$$
Introduce set $\Hcal$ satisfying property $P(\theta)$ as stated in Section 4.
Now define
$$
\Zcal=\Bigl\{z\in \mathbb R_+^{d}:
\sum_{i\in M\setminus M^\star} z_{i}\geq \frac{1}{g_M},\;\; \forall M
\in \Hcal\Big\},
$$
and
\begin{align*}
\textsf{P3:}\quad \inf_{z\in\Zcal} \beta(\theta)\sum_{i\in [d]\setminus M^\star} z_{i}.
\end{align*}
Observe that $\textrm{val}(\textsf{P2})\ge\textrm{val}(\textsf{P3})$ since the feasible set of \textsf{P2} is contained in $\Zcal$.
The definition of $\Hcal$ implies that $\sum_{i\in [d]\setminus M^\star} z_i=\sum_{M\in \Hcal} \sum_{i\in M\setminus M^\star} z_i$.
It then follows that
\begin{align*}
\textrm{val}(\textsf{P3})&= \sum_{M\in\Hcal} \frac{\beta(\theta)}{g_M}\\
                         &\ge \sum_{M\in\Hcal} \frac{\beta(\theta)}{\max_{i\in M\setminus M^\star} \klber(\theta_i,\hat\lambda_i(M))} \\
                         & =  \sum_{M\in\Hcal} \frac{\beta(\theta)}{\max_{i\in M\setminus M^\star} \klber\left(\theta_i,\frac{1}{|M\setminus M^\star|}\sum_{j\in M^\star\setminus M}\theta_j\right)}.
                       %  &\ge \frac{\beta(\theta)|\Hcal|}{\max_{M\in\Hcal}{g_M}}.
\end{align*}
The proof is completed by observing that:
$c(\theta)\ge \textrm{val}(\textsf{P1})\ge \textrm{val}(\textsf{P2})\ge \textrm{val}(\textsf{P3})$.
\ep

\subsection{Proof of Corollary 1}
Fix $M\neq M^\star$. For any $i\in M\setminus M^\star$, we have:
\begin{align*}
\klber\Bigl(\theta_i,\frac{1}{|M\setminus M^\star|}\sum_{j\in M^\star\setminus M}\theta_j\Big)
&\le \frac{1}{|M\setminus M^\star|}\sum_{j\in M^\star\setminus M} \klber\left(\theta_i,\theta_j\right) \quad \hbox{(By convexity of $\klber(.,.)$)}\\
&\le \frac{1}{|M\setminus M^\star|}\sum_{j\in M^\star\setminus M} \frac{(\theta_i-\theta_j)^2}{\theta_j(1-\theta_j)} \\
&\le \frac{1}{|M\setminus M^\star|}\sum_{j\in M^\star\setminus M} \frac{(1-\theta_j)^2}{\theta_j(1-\theta_j)} \\
&\le \frac{1}{|M\setminus M^\star|}\sum_{j\in M^\star\setminus M} \left(\frac{1}{\theta_j}-1\right)\\
&\le \frac{1}{\min_{j\in M^\star\setminus M}\theta_j}-1\\
&\le \frac{1}{a}-1,
\end{align*}
where the second inequality follows from the inequality $\klber(p,q) \leq \frac{(p-q)^2}{q(1-q)}$ for all $(p,q)  \in [0,1]^2$.
Moreover, we have that
\begin{align*}
\beta(\theta)=\min_{M\neq M^\star}\frac{\Delta^M}{|M\setminus M^\star|}\ge \frac{\Delta_{\min}}{\max_{M} |M\setminus M^\star|}=\frac{\Delta_{\min}}{k}.
\end{align*}

Applying Theorem 2, we get:
\begin{align*}
c(\theta)\ge \sum_{M\in\Hcal} \frac{\beta(\theta)}{\max_{i\in M\setminus M^\star} \klber\left(\theta_i,\frac{1}{|M\setminus
M^\star|}\sum_{j\in M^\star\setminus M}\theta_j\right)}\ge \frac{\Delta_{\min}a}{k(1-a)}|\Hcal|,
\end{align*}
which gives the required lower bound and completes the proof.
\ep

%\subsection{Proof of Corollary 1}
%Assume that $\theta =  y M^\star + x(1-M^\star)$. Hence $\mu^\star = m y$ and for all $M$: $M^\top \theta =  y (m - | M^\star/M|) + x | M^\star/M|$. Hence $\mu^\star - M^{\top} \theta = |M^\star/M|(y-x)$ so that $\beta(\theta) = y-x$. Also $\Delta_{\min} = \min_{M \in {\cal M}} |M^\star/M|(y-x) \geq a(y-x)$ since ${\cal M}$ is $a$-flippable. Hence $(y-x) \leq \Delta_{\min}/a$. Consider $i$ such that $M^\star_i = 0$, and $M \neq M^\star$. Then:
%\eqs{
%	\klber \Bigl( \theta_i, |M \setminus M^\star|^{-1} \sum_{j \in M^\star \setminus M} \theta_j \Big) = \klber(x,y) \leq \frac{(x-y)^2}{y(1-y)}.
%}
%We have used the fact that for all $(p,q)  \in [0,1]^2$, $\klber(p,q) \leq \frac{(p-q)^2}{q(1-q)}$. This inequality comes from the fact that $\log(u) \leq u-1$, $u \geq 0$. Define $I = \{i: M^{\star}_i = 0 \}$. We have $|I|=d-m \geq a\floor{(d-m)/a}$. Define $K = \floor{(d-m)/a}$ and define $I_1,...,I_K$ disjoint subsets of $I$ with $|I_k| = a$ for all $k$. We have assumed that ${\cal M}$ is $a$-flippable, so for each $k$, there exists $M^k \in {\cal M}$ such that $I_k \subset M^k$ and $(M^k \setminus A) \subset M^\star$.Now define ${\cal H} = \{M^1,...,M^K\}$. Applying the first statement of the theorem and using the above we get:
%\eqs{
%c(\theta) \geq |{\cal H}| \beta(\theta) \frac{y(1-y)}{(x-y)^2} = \left\lfloor\frac{d-m}{a}\right\rfloor (y-x) \frac{y(1-y)}{(x-y)^2} \geq a \left\lfloor \frac{d-m}{a} \right\rfloor \frac{y(1-y)}{\Delta_{\min}},
%}
%which is the announced result.

\subsection{Examples of Scaling of the Lower Bound}

\subsubsection{Matchings}
\label{sec:matching_prb_1}
\begin{figure}
\begin{center}
\subfigure[$M^\star$]{
\includegraphics[scale=.5,angle=0]{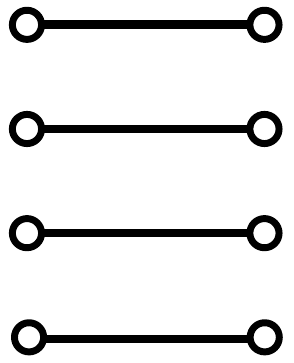}}
\subfigure[]{
\includegraphics[scale=.5,angle=0]{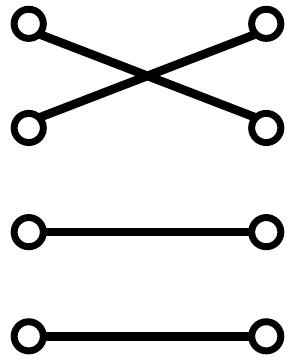}}
\subfigure[]{
\includegraphics[scale=.5,angle=0]{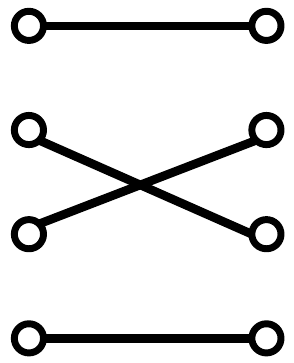}}
\subfigure[]{
\includegraphics[scale=.5,angle=0]{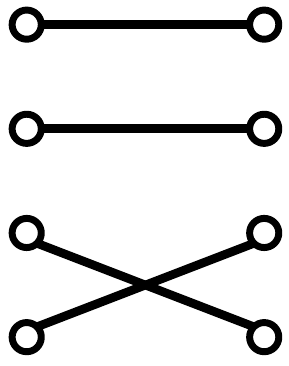}}
\subfigure[]{
\includegraphics[scale=.5,angle=0]{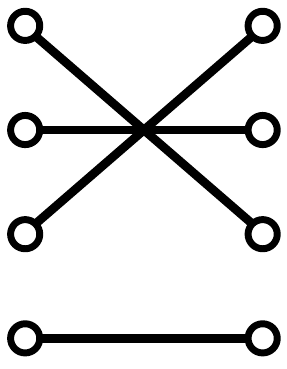}}
\subfigure[]{
\includegraphics[scale=.5,angle=0]{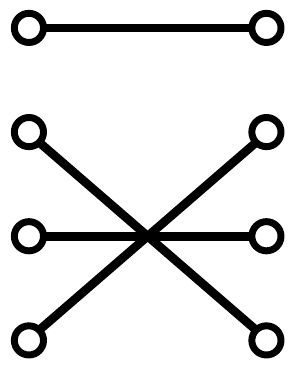}}
\subfigure[]{
\includegraphics[scale=.5,angle=0]{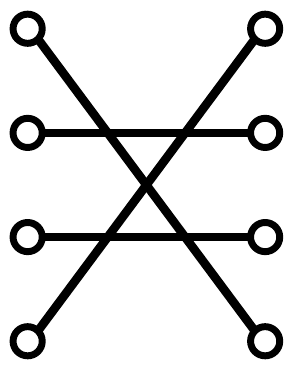}}
\end{center}
\caption{Matchings in $\Kcal_{4,4}$: (a) The optimal matching $M^\star$, (b)-(g) Elements of $\Hcal$.}
\label{fig:Hcal_matching_n4c4}
\end{figure}

In the first example, we assume that $\Mcal$ is the set of perfect matchings in the complete bipartite graph $\Kcal_{m,m}$, with $|\Mcal|=m!$ and $d=m^2$. A maximal subset ${\cal H}$ of ${\cal M}$ satisfying property $P(\theta)$ can be constructed by adding all matchings that differ from the optimal matching by only two edges, see Figure \ref{fig:Hcal_matching_n4c4} for illustration in the case of $m=4$. Here $|{\cal H}| = {m\choose 2}$ and thus, $|\Hcal|$ scales as $m^2=d$.

\subsubsection{Spanning trees}
\label{sec:spanning_tree_prb_1}
\begin{figure}
\begin{center}
\subfigure[$M^\star$]{
\includegraphics[scale=.5,angle=0]{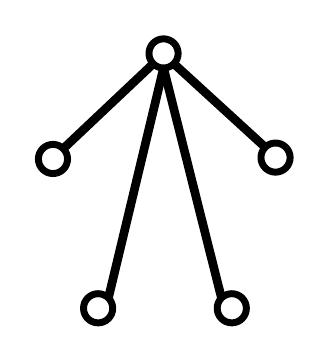}}
\subfigure[]{
\includegraphics[scale=.5,angle=0]{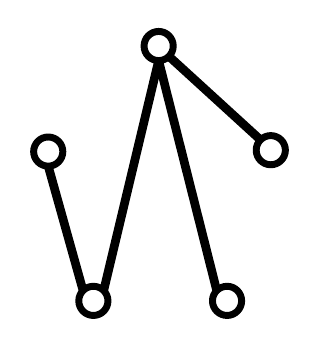}}
\subfigure[]{
\includegraphics[scale=.5,angle=0]{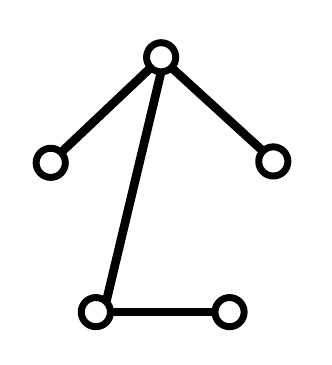}}
\subfigure[]{
\includegraphics[scale=.5,angle=0]{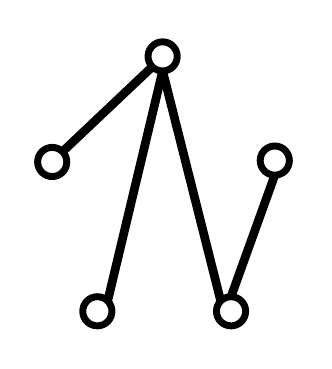}}
\subfigure[]{
\includegraphics[scale=.5,angle=0]{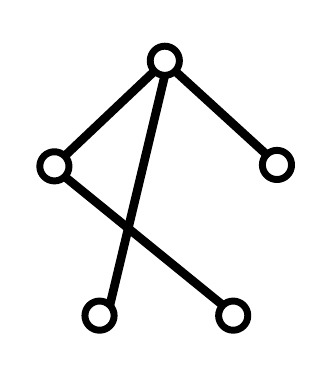}}
\subfigure[]{
\includegraphics[scale=.5,angle=0]{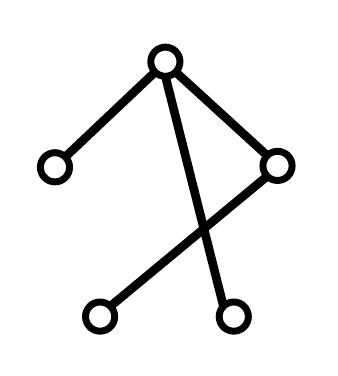}}
\subfigure[]{
\includegraphics[scale=.5,angle=0]{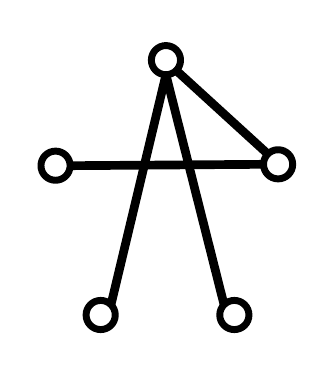}}
\end{center}
\caption{Spanning trees in $\Kcal_{5}$: (a) The optimal spanning tree $M^\star$, (b)-(g) Elements of $\Hcal$.}
\label{fig:Hcal_MST}
\end{figure}

Consider the problem of finding the minimum spanning tree in a complete graph $\Kcal_{N}$. This corresponds to letting $\Mcal$  be the set of all spanning trees in $\Kcal_N$, where $|\Mcal|=N^{N-2}$ (Cayley's formula). In this case, we have $d={N\choose 2}=\frac{N(N-1)}{2}$, which is the number of edges of $\Kcal_N$, and $m=N-1$. A maximal subset ${\cal H}$ of ${\cal M}$ satisfying property $P(\theta)$ can be constructed by composing all spanning trees that differ from the optimal tree by one edge only, see Figure \ref{fig:Hcal_MST}. In this case, $\Hcal$ has $d-m=\frac{(N-1)(N-2)}{2}$ elements.

\subsubsection{Routing in a grid}
\label{sec:routing_grid}
\begin{figure}
\begin{center}
\subfigure[]{
\includegraphics[scale=.38,angle=0]{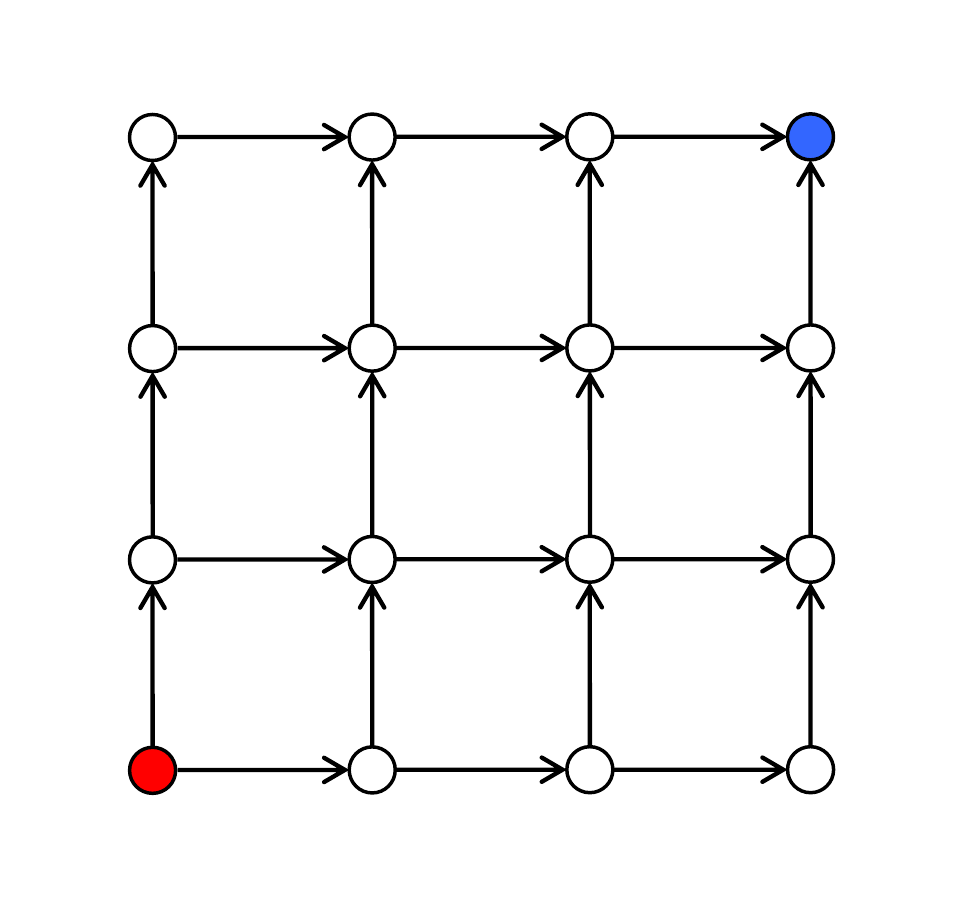}}
\subfigure[]{
\includegraphics[scale=.38,angle=0]{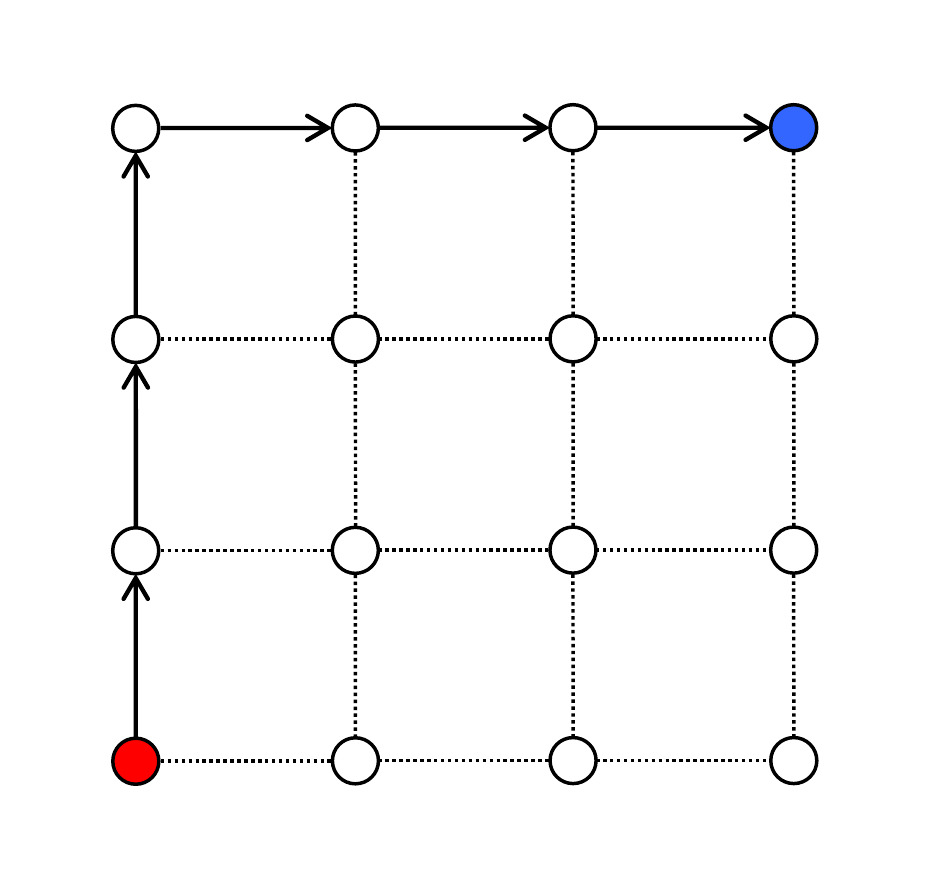}}
\subfigure[]{
\includegraphics[scale=.38,angle=0]{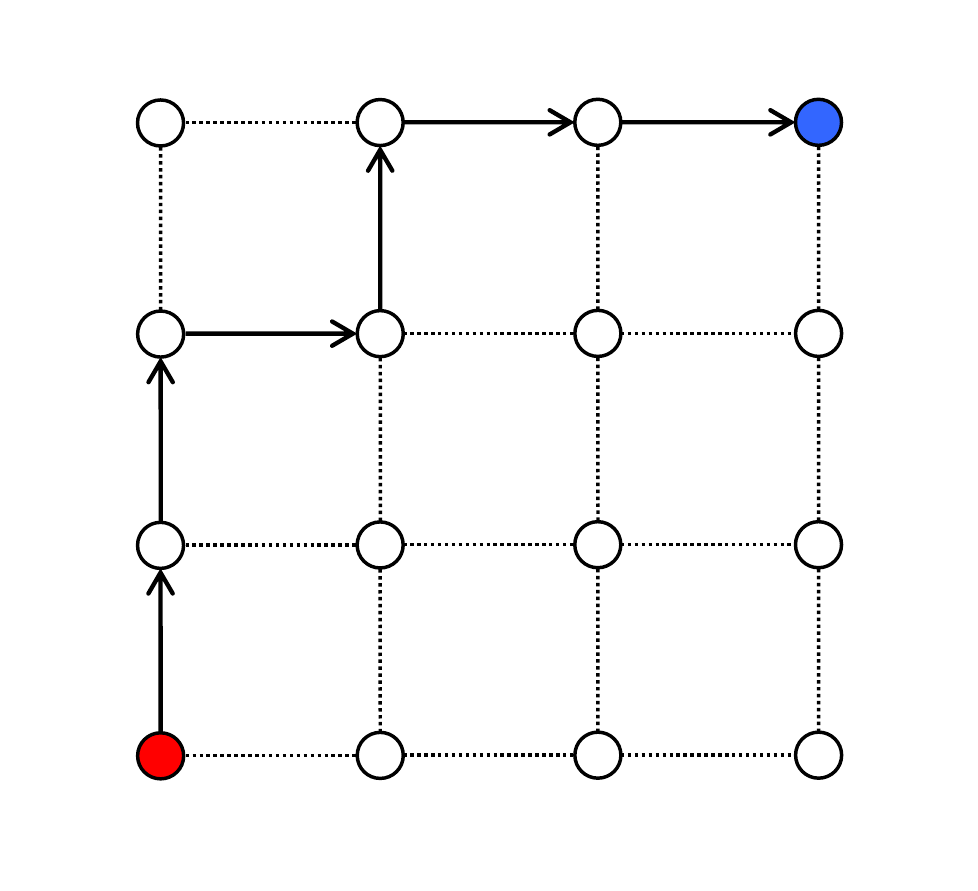}}
\subfigure[]{
\includegraphics[scale=.38,angle=0]{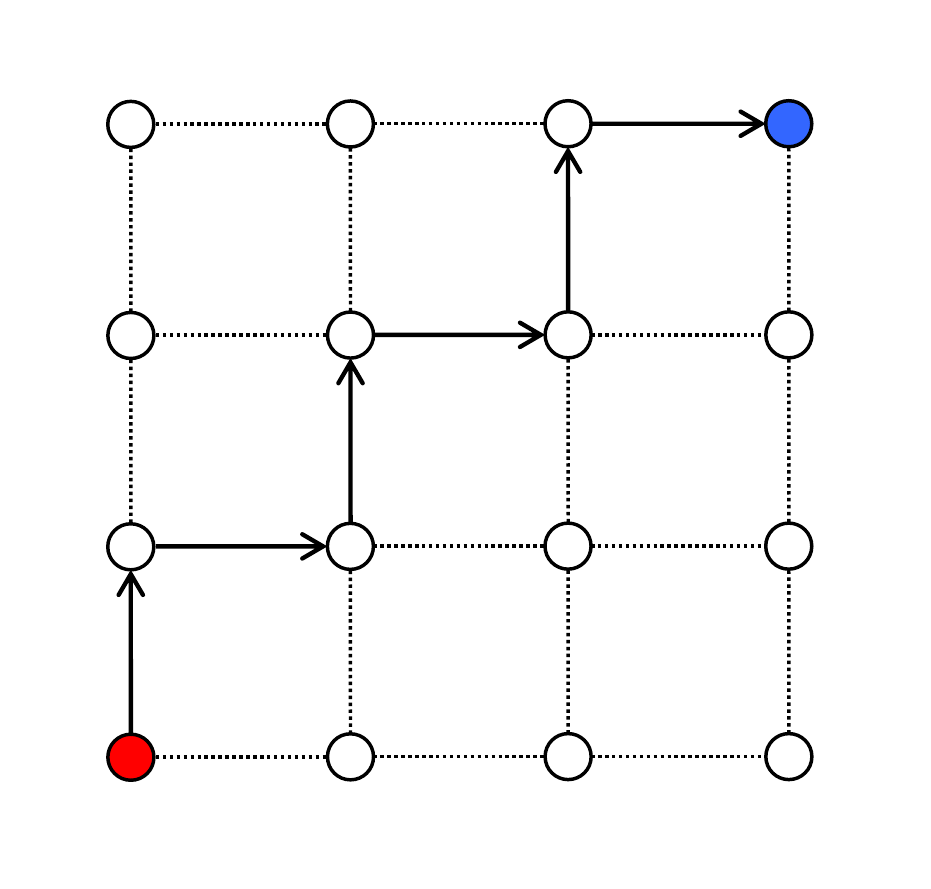}}
\subfigure[]{
\includegraphics[scale=.38,angle=0]{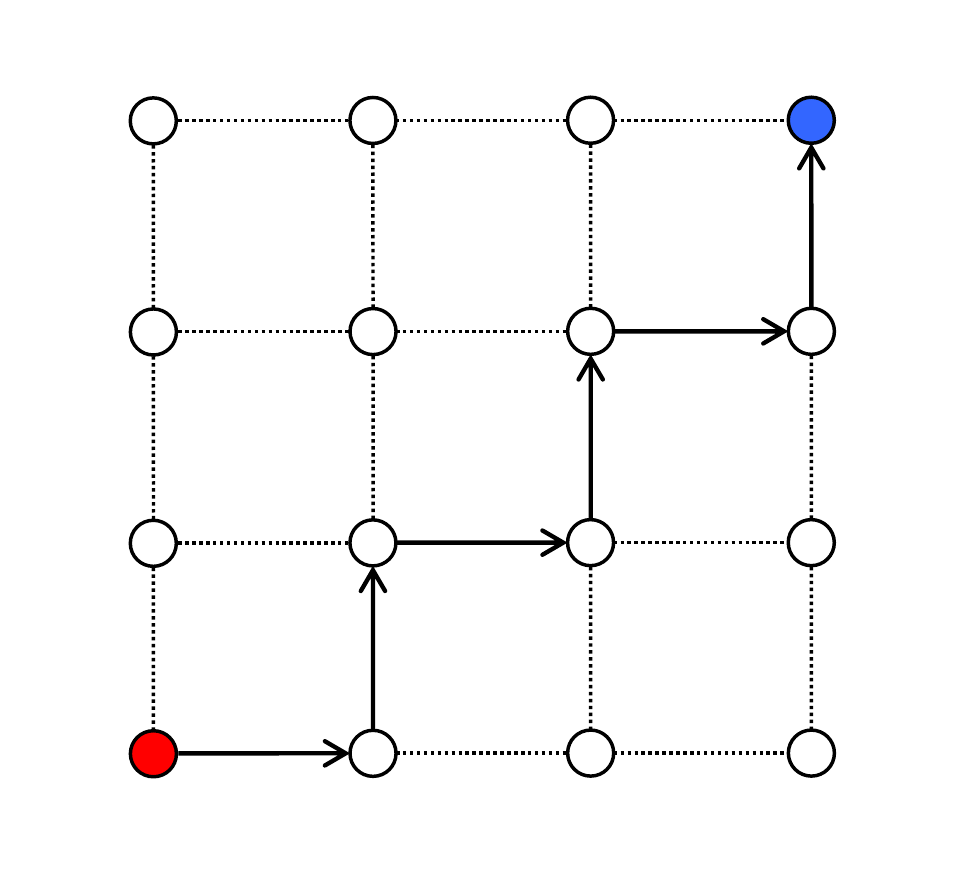}}
\end{center}
\caption{Routing in a grid: (a) Grid topology with source (red) and destination (blue) nodes, (b) Optimal path $M^\star$, (c)-(e) Elements of $\Hcal$.}
\label{fig:Hcal_routing}
\end{figure}

Now we give an example, in which $|\Hcal|$ is not scaling as $\Omega(d)$. Consider routing in an $N$-by-$N$ directed grid, whose topology is shown in Figure \ref{fig:Hcal_routing}(a) where the source (resp. destination) node is shown in red (resp. blue). Here $\Mcal$ is the set of all ${{2N-2}\choose{N-1}}$ paths with $m = 2(N-1)$ edges. We further have $d = 2N(N-1)$.
In this example, elements of any maximal set $\Hcal$ satisfying $P(\theta)$ do not cover all basic actions. For instance, for the grid shown in Figure \ref{fig:Hcal_routing}(a), the two edges incident to the right lower corner do not appear in any arm in $\Hcal$. It can be easily verified that in this case, $|\Hcal|$ scales as $N$ rather than $N^2=d$.

\subsection{Lower Bound Example}

\begin{figure}[!th]
\begin{center}
\includegraphics[scale=.5]{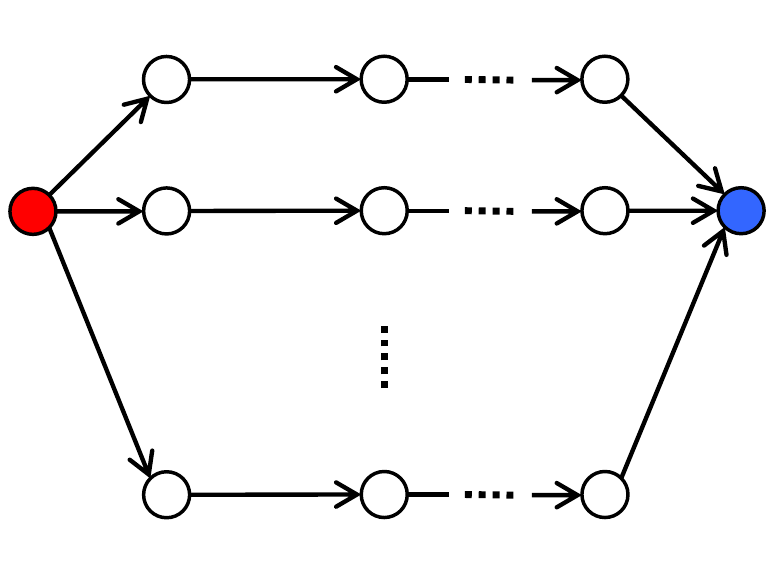}
\end{center}
\caption{Lower bound example}
\label{fig:LB_topology}
\end{figure}

Here we provide an example, motivated by \cite{kveton2014tight}, to investigate the tightness of the regret bounds of our algorithms. Consider the topology shown in Figure \ref{fig:LB_topology}, where there are $\frac{d}{m}$ paths, each consisting of $m$ links. Let parameter $\theta$ be defined such that
\begin{align*}
\theta_i=
\begin{cases}
  0.5 & \; \hbox{if $i$ belongs to the first path} \\
  0.5-\delta  & \;  \hbox{otherwise.}
  \end{cases}
\end{align*}
The first path is the optimal path and for any $M\neq 1$ we have: $\Delta_M=\Delta=m\delta.$
Since various paths are independent, this problem reduces to a classical MAB problem with $\frac{d}{m}$ arms. It is observed that the total reward of each path is the sum of $m$ independent Bernoulli random variables with the same parameter. Hence, it is distributed according to a binomial distribution. It then follows that

\begin{align*}
\liminf_{T\to \infty} \frac{R(T)}{\log(T)}
&\ge \sum_{M\neq M^\star} \frac{\Delta_M}{\kl(\mathrm{Bin}(m,0.5-\delta),\mathrm{Bin}(m,0.5))} \\
&= \left(\frac{d}{m}-1\right)\cdot \frac{\Delta}{m\klber(0.5-\delta,0.5)} \\
&\ge \frac{(d-m)\Delta}{4m^2\delta^2} \\
&= \frac{d-m}{4\Delta},
\end{align*}
where the first equality follows from the fact that the KL divergence between two Binomial distributions with respective parameters $(m,u)$ and $(m,v)$ is $m \klber(u,v)$, and where the last step is due to inequality $\klber(x,y)\le \frac{(x-y)^2}{y(1-y)}$ for all $x,y\in (0,1)$.

\section{Stochastic Combinatorial Bandits: Regret Analysis of~\algos}
\label{sec:supp_stochastic}

We use the convention that for $v,u \in \RR^d$, $vu = (v_i u_i)_{i \in [d]}$.

\subsection{A concentration inequality}
We first recall Lemma~\ref{lem:colt}, a concentration inequality derived in \cite[Theorem 2]{magureanu2014lipschitz}.

\begin{lemma}\label{lem:colt}
	There exists a number $C_m > 0$ depending only on $m$ such that, for all $M$ and all $n\ge 2$:
	\eqs{
	\PP[ (M t(n))^\top \klber(\hat\theta(n),\theta) \geq f(n)] \leq C_m n^{-1} (\log(n))^{-2}.
	}
\end{lemma}

\subsection{Proof of Theorem 3}

\underline{\textbf{First statement:}}

 Consider $q \in \Theta$, and apply the Cauchy-Schwartz inequality:
	\eqs{
	 M^\top (q - \hat\theta(n)) =  \sum_{i=1}^d  \sqrt{t_i(n)} (q_i - \hat\theta_i(n))  \frac{M_i}{\sqrt{t_i(n)}}
	 \leq \sqrt{ \sum_{i=1}^d M_i t_i(n) (q_i - \hat\theta_i(n))^2 }\sqrt{ \sum_{i=1}^d\frac{M_i}{t_i(n)}}
	}
	By Pinsker's inequality, for all $(p,q) \in [0,1]^2$ we have $2 (p-q)^2 \leq \klber(p,q)$ so that:
	\eqs{
	M^\top (q - \hat\theta(n))  \leq \sqrt{ \frac{(M t(n))^\top \klber(\hat\theta(n),q)}{2}}\sqrt{\sum_{i=1}^d \frac{M_i}{t_i(n)}}
	}
	Hence, $(M t(n))^\top \klber(\hat\theta(n),q) \leq f(n)$ implies:
	\eqs{
		M^\top q = M^\top \hat\theta(n) + M^\top (q - \hat\theta(n)) \leq M^\top \hat\theta(n) + \sqrt{ \frac{f(n)}{2} \sum_{i=1}^d \frac{M_i}{t_i(n)} } = c_M(n).
	}
	so that, by definition of $b_M(n)$, we have $b_M(n) \leq c_M(n)$.	

\underline{\textbf{Second statement:}}

	If $(M t(n))^\top \klber(\hat\theta(n),\theta) \leq f(n)$ then, by definition of $b_M(n)$ we have $b_M(n) \geq M^\top \theta$. Therefore, using Lemma~\ref{lem:colt}, there exists $C_m$ such that for all $n\ge 2$ we have:
	\eqs{
		\PP[b_M(n) < M^\top \theta] \leq \PP[(M t(n))^\top \klber(\hat\theta(n),\theta) \geq f(n)] \leq C_m n^{-1} (\log(n))^{-2},
	}
which concludes the proof.

\subsection{Proof of Theorem 4}

We recall the following facts about the KL divergence $\klber$, for all $p \in [0,1]$:
\begin{itemize}
\item[(i)]  $q \mapsto \klber(p,q)$ is strictly convex on $[0,1]$ and attains its minimum at $p$, with $\klber(p,p) = 0$.
\item[(ii)] Its derivative with respect to the second parameter $q \mapsto \klber'(p,q) = \frac{q-p}{q(1-q)}$ is strictly increasing on $(p,1)$.
\item[(iii)] For $p < 1$, we have $\klber(p,q) \tto{q \to 1^{-}} \infty$ and $\klber'(p,q) \tto{q \to 1^{-}} \infty$.
\end{itemize}

Consider $M$ and $n$ fixed throughout the proof. Define $I = \{i \in M: \hat\theta_i(n) \neq 1\}$. Consider $q^\star \in \Theta$ the optimal solution of optimization problem:
\als{
	 	\max_{q \in \Theta} \; & M^\top q \sk
		\text{s.t. } & (Mt(n))^\top \klber(\hat\theta(n),q) \leq f(n).
}
so that $b_M(n) = M^\top q^\star$. Consider $i \not\in M$, then $M^\top q$ does not depend on $q_i$ and from (i) we get $q_i = \hat\theta_i(n)$. Now consider $i \in M$. From (i) we get that $1 \geq q^\star_i \geq \hat\theta_i(n)$. Hence $q^\star_i = 1$ if $\hat\theta_i(n) = 1$. If $I$ is empty, then $q^\star_i = 1$ for all $i \in M$, so that $b_M(n) = || M ||_1$.

Consider the case where $I \neq \emptyset$. From (iii) and the fact that $t(n)^\top \klber(\hat\theta(n),q^\star) < \infty$ we get $\hat\theta_i(n) \leq q^\star_i < 1$. From the Karush-Kuhn-Tucker (KKT) conditions, there exists $\lambda^\star > 0$ such that for all $i \in I$:
\eqs{
1 = \lambda^\star t_i(n) \klber'(\hat\theta_i(n),q^\star_i).
}
For $\lambda > 0$ define $\hat\theta_i(n) \leq \overline{q}_i(\lambda) < 1$ a solution to the equation:
\eqs{
1 = \lambda t_i(n) \klber'(\hat\theta_i(n),\overline{q}_i(\lambda)).
}
From (i) we have that $\lambda \mapsto \overline{q}_i(\lambda)$ is uniquely defined, is strictly decreasing and $\hat\theta_i(n) < \overline{q}_i(\lambda) < 1$. From (iii) we get that $\overline{q}_i(\RR^+) = [ \hat\theta_i(n),1]$. Define the function:
\eqs{
F(\lambda) = \sum_{i \in I} t_i(n) \klber(\hat\theta(n),\overline{q}_i(\lambda)).
}
From the reasoning below, $F$ is well defined, strictly increasing and $F(\RR^+) = \RR^{+}$. Therefore, $\lambda^\star$ is the unique solution to $F(\lambda^\star) = f(n)$, and $q_i^\star = \overline{q}_i(\lambda^\star)$. Furthermore, replacing $\klber'$ by its expression we obtain the quadratic equation:
\eqs{
\overline{q}_i(\lambda)^2 + \overline{q}_i(\lambda)( \lambda t_i(n) - 1) - \lambda t_i(n) \hat\theta_i(n) = 0.
}
Solving for $\overline{q}_i(\lambda)$, we obtain that $\overline{q}_i(\lambda) = g(\lambda,\hat\theta_i(n),t_i(n))$, which concludes the proof.
\ep

\subsection{Proof of Theorem 5}
To prove Theorem 5, we borrow some ideas from proof of \cite[Theorem~3]{kveton2014tight}.

For any $n\in \mathbb{N}$, $s\in \mathbb{R}^d$, and $M\in\Mcal$ define $h_{n,s,M}=\sqrt{\frac{f(n)}{2}\sum_{i=1}^d \frac{M_i}{s_i}}$, and
introduce the following events:
\begin{align*}
G_{n}&=\{(M^\star t(n))^\top \klber(\hat\theta(n), \theta)> f(n)\},\\
H_{i,n}&=\{M_i(n)=1,\; |\hat\theta_i(n)-\theta_i|\ge m^{-1}\Delta_{\min}/2\},\;\; H_n=\cup_{i=1}^d H_{i,n},\\
F_n&=\{\Delta_{M(n)} \le 2h_{T,t(n),M(n)}\}.
\end{align*}
Then the regret can be bounded as:
\begin{align*}
R^\pi(T)&= \EE [\sum_{n=1}^T \Delta_{M(n)}]\le \EE [\sum_{n=1}^T \Delta_{M(n)}(\mathbbmss 1\{G_n\}+ \mathbbmss 1\{H_n\})] +
\EE [\sum_{n=1}^T \Delta_{M(n)} \mathbbmss 1 \{\overline{G_n}, \; \overline{H_n}\}]\\
&\le m\EE [\sum_{n=1}^T (\mathbbmss 1\{G_n\}+\mathbbmss 1\{H_n\})]
+ \EE [\sum_{n=1}^T \Delta_{M(n)}\mathbbmss 1\{\overline{G_n}, \;\overline{H_n}\}],
\end{align*}
since $\Delta_{M(n)}\le m$.

Next we show that for any $n$ such that $M(n)\neq M^\star$, it holds that $\overline{G_n\cup H_n} \subset F_n$.
Recall that $c_M(n)\ge b_M(n)$ for any $M$ and $n$ (Theorem 3). Moreover, if $\overline{G_n}$ holds, we have $(M^\star t(n))^\top \klber(\hat\theta(n), \theta)\le f(n)$, which by definition of $b_M$ implies: $b_{M^\star}(n)\ge {M^\star}^\top\theta$. Hence we have:
\begin{align*}
\mathbbmss 1\{\overline{G_n}, \; \overline{H_n}, \; M(n)\neq M^\star\}
&=\mathbbmss 1\{\overline{G_n}, \; \overline{H_n}, \; \xi_{M(n)}(n)\ge \xi_{M^\star}(n)\}\\
%&=\Delta_{M(n)}\mathbbmss 1\{\overline{G_n}, \xi_{M(n)}(n)\ge \xi_{M^\star}(n)\} %\mathbbmss 1\{ (M^\star t(n))^\top \klber(\hat\theta(n), \theta)\le f(n)\}\\
&\le \mathbbmss 1\{\overline{H_n}, \; c_{M(n)}(n)\ge {M^\star}^\top\theta\}\\ %\mathbbmss 1\{ (M^\star t(n))^\top \klber(\hat\theta(n), \theta)\le f(n)\}\\
%&\le \Delta_{M(n)}\mathbbmss 1\{\overline{G_n}, c_{M(n)}(n)\ge {M^\star}^\top\theta\} \\
&= \mathbbmss 1\{\overline{H_n}, \; M(n)^\top\hat\theta(n)+h_{n,t(n),M(n)}\ge {M^\star}^\top\theta\}\\
&\le  \mathbbmss 1 \{M(n)^\top\theta+ \Delta_{M(n)}/2 +h_{n,t(n),M(n)}\ge {M^\star}^\top\theta \}\\
&=  \mathbbmss 1 \{2h_{n,t(n),M(n)}\ge \Delta_{M(n)}\}\\
&\le  \mathbbmss 1 \{2h_{T,t(n),M(n)}\ge \Delta_{M(n)}\}\\
&=  \mathbbmss 1 \{F_n\},
\end{align*}
where the second inequality follows from the fact that event $\overline{G_n}$ implies: $M(n)^\top\hat\theta(n)\le M(n)^\top\theta+\Delta_{\min}/2\le M(n)^\top\theta +\Delta_{M(n)}/2$.

Hence, the regret is upper bounded by:
\begin{align*}
R^\pi(T)&\le m \EE[\sum_{n=1}^T \mathbbmss 1\{G_n\}]+m \EE[\sum_{n=1}^T \mathbbmss 1\{H_n\}]
+ \EE [\sum_{n=1}^T \Delta_{M(n)}\mathbbmss 1\{F_n\}].
\end{align*}

We will prove the following inequalities: (i) $\EE[\sum_{n=1}^T \mathbbmss 1\{G_n\}] \le m^{-1}C'_m,$ with $C'_m\ge 0$ independent of $\theta$, $d$, and $T$, (ii) $\EE[\sum_{n=1}^T \mathbbmss 1\{H_n\}]\le 4dm^2\Delta_{\min}^{-2}$,
and (iii) $\EE [\sum_{n=1}^T \Delta_{M(n)}\mathbbmss 1\{F_n\}]\le 16d\sqrt{m}\Delta_{\min}^{-1}f(T)$.

Hence as announced:
\begin{align*}
R^\pi(T)&\le 16d\sqrt{m}\Delta_{\min}^{-1}f(T) + 4dm^3\Delta_{\min}^{-2} + C'_m.
\end{align*}

\noindent\underline{\textbf{Inequality (i):}} An application of Lemma \ref{lem:colt} gives
\begin{align*}
\EE[\sum_{n=1}^T \mathbbmss 1\{G_n\}]&=\sum_{n=1}^T \PP[(M^\star t(n))^\top \klber(\hat\theta(n), \theta)> f(n)]\\
& \le 1+\sum_{n\ge 2}C_m n^{-1}(\log(n))^{-2}\equiv m^{-1}C'_{m}<\infty.
\end{align*}

\noindent\underline{\textbf{Inequality (ii):}} 
Fix $i$ and $n$. Define $s = \sum_{n'=1}^n \indic\{H_{n',i}\}$. Observe that $H_{n',i}$ implies $M_i(n')=1$, hence $t_i(n) \geq s$. Therefore, applying \cite[Lemma~B.1]{combes2014unimodal_techreport}, we have that $\sum_{n=1}^T \PP[H_{n,i}] \leq 4m^2\Delta_{\min}^{-2}$.
Using the union bound: $\sum_{n=1}^T \PP[H_{n}] \leq 4dm^2\Delta_{\min}^{-2}.$

\noindent\underline{\textbf{Inequality (iii):}} Let $\ell>0$. For any $n$ introduce the following events:
\begin{align*}
S_n&=\{i\in M(n) : t_i(n) \le 4 mf(T)\Delta_{M(n)}^{-2}\},\\
A_{n}&=\{|S_n|\ge \ell\},\\
B_{n}&=\{|S_n|< \ell, \;[\exists i\in M(n): t_i(n)\le 4\ell f(T)\Delta_{M(n)}^{-2}]\}.
\end{align*}
We claim that for any $n$ such that $M(n)\neq M^\star$, we have $F_n \subset (A_n \cup B_n)$. To prove this, we show that when $F_n$ holds and $M(n)\neq M^\star$, the event $\overline{A_{n}\cup B_{n}}$ cannot happen.
Let $n$ be a time instant such that $M(n)\neq M^\star$ and $F_n$ holds, and assume that $\overline{A_n\cup B_n}=\{|S_n|<\ell, \;[\forall i\in M(n): t_i(n)> 4\ell f(T)\Delta_{M(n)}^{-2}]\}$ happens. Then $F_n$ implies:
\begin{align}
\Delta_{M(n)}&\le 2h_{T,t(n),M(n)}=2\sqrt{\frac{f(T)}{2}} \sqrt{\sum_{i\in [d]\setminus S_n} \frac{M_i(n)}{t_i(n)}+ \sum_{i\in S_n} \frac{M_i(n)}{t_i(n)}} \nonumber\\
\label{F_n_contrdic}
            &<2\sqrt{\frac{f(T)}{2}} \sqrt{m \frac{\Delta_{M(n)}^2}{4 mf(T)}+ |S_n| \frac{\Delta_{M(n)}^2}{4\ell f(T)}}<\Delta_{M(n)},%\sqrt{\frac{1}{2}+ \ell \frac{1}{2\ell}}=\Delta_{M(n)},
\end{align}
where the last inequality uses the observation that $\overline{A_n\cup B_n}$ implies  $|S_n|<\ell$. Clearly, (\ref{F_n_contrdic}) is a contradiction. Thus $F_n \subset (A_n \cup B_n)$ and consequently:
%Hence either $A_n$ or $B_n$ happens and thus
\begin{align}
\label{eq:F_n_regret}
\sum_{n=1}^T \Delta_{M(n)}\mathbbmss 1\{F_n\}&\le \sum_{n=1}^T \Delta_{M(n)}\mathbbmss 1\{A_n\} + \sum_{n=1}^T \Delta_{M(n)}\mathbbmss 1\{B_n\}.
\end{align}
To further bound the r.h.s. of the above, we introduce the following events for any $i$:
\begin{align*}
A_{i,n}&=A_n\cap \{i\in M(n), \;t_i(n)\le 4 mf(T)\Delta_{M(n)}^{-2}\},\\
B_{i,n}&=B_n\cap \{i\in M(n), \;t_i(n)\le 4 \ell f(T)\Delta_{M(n)}^{-2}\}.
\end{align*}
It is noted that: %$A_{i,n}=\{|S_n|\ge \ell,\; i\in S_n\}$, and hence
$$
\sum_{i\in [d]} \mathbbmss 1 \{A_{i,n}\}=\mathbbmss 1 \{A_n\} \sum_{i\in [d]} \mathbbmss 1 \{i\in S_n\}= |S_n|\mathbbmss 1 \{A_n\}\ge \ell \mathbbmss 1 \{A_n\},
$$
and hence: $\mathbbmss 1 \{A_n \}\le \frac{1}{\ell}\sum_{i\in [d]} \mathbbmss 1 \{A_{i,n}\}$. Moreover $\mathbbmss 1 \{B_n \}\le \sum_{i\in [d]} \mathbbmss 1 \{B_{i,n}\}$.
Let each basic action $i$ belong to $K_i$ suboptimal arms, ordered based on their gaps as: $\Delta^{i,1}\ge \dots \ge \Delta^{i,K_i}>0$. Also define $\Delta^{i,0}=\infty$.
%Also let $D\subseteq [d]$ be set containing basic actions that belong to at least a suboptimal arm (i.e., for any $i\in D$, there exists $M\neq M^\star$ such that $M_i=1$).
Plugging the above inequalities into (\ref{eq:F_n_regret}), we have
\begin{align*}
\sum_{n=1}^T \Delta_{M(n)}\mathbbmss 1\{F_n\}&\le \sum_{n=1}^T\sum_{i=1}^d \frac{\Delta_{M(n)}}{\ell} \mathbbmss 1 \{A_{i,n}\} + \sum_{n=1}^T \sum_{i=1}^d \Delta_{M(n)}\mathbbmss 1 \{B_{i,n}\}\\
&= \sum_{n=1}^T\sum_{i=1}^d \frac{\Delta_{M(n)}}{\ell} \mathbbmss 1 \{A_{i,n}, \; M(n)\neq M^\star \} + \sum_{n=1}^T \sum_{i=1}^d \Delta_{M(n)}\mathbbmss 1 \{B_{i,n}, \; M(n)\neq M^\star \}\\
&\le \sum_{n=1}^T \sum_{i=1}^d \sum_{k\in [K_i]}\frac{\Delta^{i,k}}{\ell}\mathbbmss 1\{A_{i,n}, \; M(n)=k \} + \sum_{n=1}^T\sum_{i=1}^d \sum_{k\in [K_i]} \Delta^{i,k}\mathbbmss 1\{B_{i,n}, \; M(n)=k \}\\
    &\le \sum_{i=1}^d \sum_{n=1}^T \sum_{k\in [K_i]}\frac{\Delta^{i,k}}{\ell}\mathbbmss 1\{i\in M(n), \;t_i(n)\le 4 mf(T)(\Delta^{i,k})^{-2}, \; M(n)=k \}\\
    &+ \sum_{i=1}^d \sum_{n=1}^T \sum_{k\in [K_i]} \Delta^{i,k}\mathbbmss 1\{i\in M(n), \;t_i(n)\le 4 \ell f(T)(\Delta^{i,k})^{-2}, \; M(n)=k \}\\
    %&\le \sum_{i=1}^d \frac{4\alpha mf(T)}{\ell \Delta^{i,K_i}}
    %+ \sum_{i=1}^d \frac{4\alpha \ell f(T)}{(\alpha-1)\Delta^{i,K_i}}\\
    &\le \frac{8df(T)}{\Delta_{\min}}\left(\frac{m}{\ell}+\ell\right),
\end{align*}
where the last inequality follows from Lemma \ref{lem:regret_supp1}, which is proven next. The proof is completed by setting $\ell=\sqrt{m}$.
\ep

\begin{lemma}
\label{lem:regret_supp1}
Let $C>0$ be a constant independent of $n$. Then for any $i$ such that $K_i\ge 1$:
\begin{align*}
\sum_{n=1}^T\sum_{k=1}^{K_i} \mathbbmss 1 \{i\in M(n), \; t_i(n)\le C(\Delta^{i,k})^{-2}, \; M(n)=k \}\Delta^{i,k} \le \frac{2C}{\Delta_{\min}}.
\end{align*}
\end{lemma}

\bp
We have:
\begin{align*}
\sum_{n=1}^T\sum_{k=1}^{K_i} \mathbbmss 1 &\{i\in M(n), \; t_i(n)\le C(\Delta^{i,k})^{-2}, \; M(n)=k \}\Delta^{i,k}\\
&= \sum_{n=1}^T\sum_{k=1}^{K_i} \sum_{j=1}^k \mathbbmss 1\{i\in M(n),\; t_i(n)\in (C(\Delta^{i,j-1})^{-2},C(\Delta^{i,j})^{-2}],\; M(n)=k \}\Delta^{i,k}\\
&\le \sum_{n=1}^T\sum_{k=1}^{K_i} \sum_{j=1}^k \mathbbmss 1 \{i\in M(n), \; t_i(n)\in ( C(\Delta^{i,j-1})^{-2}, C(\Delta^{i,j})^{-2}], \;M(n)=k \}\Delta^{i,j}\\
&\le \sum_{n=1}^T\sum_{k=1}^{K_i} \sum_{j=1}^{K_i} \mathbbmss 1 \{i\in M(n), \; t_i(n)\in ( C(\Delta^{i,j-1})^{-2}, C(\Delta^{i,j})^{-2}], \;M(n)=k \}\Delta^{i,j}\\
&\le \sum_{n=1}^T\sum_{j=1}^{K_i} \mathbbmss 1 \{i\in M(n), \; t_i(n)\in (C(\Delta^{i,j-1})^{-2}, C(\Delta^{i,j})^{-2}], \; M(n)\neq M^\star \}\Delta^{i,j}\\
&\le \frac{C}{\Delta^{i,1}} +\sum_{j=2}^{K_i} C( (\Delta^{i,j})^{-2}-(\Delta^{i,j-1})^{-2})\Delta^{i,j} \\
&\le \frac{C}{\Delta^{i,1}} + \int_{\Delta^{i,K_i}}^{\Delta^{i,2}}Cx^{-2}\mathrm{d}x\le \frac{2C}{\Delta^{i,K_i}} \le \frac{2C}{\Delta_{\min}},
\end{align*}
which completes the proof.
\ep

\subsection{\textsc{Epoch-}\algos: An algorithm with lower computational complexity}

\algos~with time horizon $T$ has a complexity of $\Ocal(|\Mcal|T)$ as neither $b_M$ nor $c_M$ can be written as $M^\top y$ for some vector $y\in\mathbb R^d$. Since $\Mcal$ typically has exponentially many elements, we deduce that \algos~is not computationally efficient. Assuming that the offline (static) combinatorial problem is solvable in $\Ocal(\OPT(\Mcal))$ time, the complexity of CUCB algorithm in \cite{chen2013combinatorial_icml} and \cite{kveton2014tight} after $T$ rounds is $\Ocal(\OPT(\Mcal) T)$. Thus, if the offline problem is efficiently implementable, i.e., $\OPT(\Mcal)=\Ocal(\mathrm{poly}(d))$, CUCB is efficient, whereas~\algos~is not. We next propose an extension to \algos, called \textsc{Epoch-}\algos, that attains almost the same regret as \algos~while enjoying much better computational complexity.

\textsc{Epoch-}\algos~algorithm in epochs of varying lengths. Epoch $k$ comprises rounds $\{N_k, \dots,N_{k+1}-1\}$, where  $N_{k+1}$ (and thus the length of the $k$-th epoch) is determined at time $n=N_k$. The algorithm simply consists in playing the arm with the maximal index at the beginning of every epoch, and playing the current leader (i.e., the arm with the highest empirical average reward) in the rest of rounds. If the leader is the arm with the maximal index, the length of epoch $k$ will be set twice as long as the previous epoch $k-1$, i.e., $N_{k+1}=N_{k}+2(N_k-N_{k-1})$. Otherwise, it will be set to 1.
In contrast to \algos, \textsc{Epoch-}\algos~computes the maximal index infrequently, and more precisely  (almost) at an exponentially decreasing rate. Thus, one might expect that after $T$ rounds, the maximal index will be computed $\Ocal(\log(T))$ times.
The pseudo-code of \textsc{Epoch-}\algos~is presented in Algorithm \ref{alg:EpochCombUCB}. %To simplify the presentation, we assume that the optimal arm $M^\star$ is unique.

\begin{algorithm}[tbh]
\small
   \caption{\textsc{Epoch-}\algos}
   \label{alg:EpochCombUCB}
\begin{algorithmic}
   \STATE {\bf Initialization:} Set $k=1$ and $N_0=N_1=1$. \vspace{1mm}
   \FOR{$n\geq 1$}
   \STATE Compute $L(n)\in \arg\max_{M\in {\cal M}} M^\top\hat\theta(n)$. %\vspace{1mm}
   \IF{$n=N_k$}
    \STATE Select arm $M(n)\in \arg\max_{M\in {\cal M}} \xi_M(n)$.
    \IF{$M(n)=L(n)$}
    \STATE Set $N_{k+1}= N_k +2(N_k-N_{k-1})$.
   \ELSE
    \STATE Set $N_{k+1}= N_k +1$.
   \ENDIF
   \STATE Increment $k$.
   \ELSE
    \STATE Select arm $M(n)=L(n)$.
   \ENDIF
   \STATE Observe the rewards, and update $t_i(n)$ and $\hat\theta_i(n), \forall i\in M(n)$. %\vspace{1mm}
   \ENDFOR
\end{algorithmic}
\normalsize
\end{algorithm}

We assess the performance of \textsc{Epoch-}\algos~through numerical experiments in the next subsection, and leave the analysis of its regret as a future work. These experiments corroborate our conjecture that he complexity of~\textsc{Epoch-}\algos~after $T$ rounds will be ${\cal O}(V(\Mcal)T+\log(T)|\Mcal|)$. Compared to CUCB, the complexity is penalized by $|\Mcal|\log(T)$, which may become dominated by the term $V(\Mcal)T$ as $T$ grows large.

\subsection{Numerical Experiments}
\label{sec:numerics}
In this section, we compare the performance of \algos~against existing algorithms through numerical experiments for some classes of $\Mcal$. When implementing \algos~we replace $f(n)$ by $\log(n)$, ignoring the term proportional to $\log(\log(n))$, as is done when implementing KL-UCB in practice.

\subsubsection{Experiment 1: Matching}
In our first experiment, we consider the matching problem with $N_1=N_2=5$, which corresponds to $d=5^2=25$ and $m=5$.
We also set $\theta$ such that $\theta_i=a$ if $i\in M^\star$, and $\theta_i=b$ otherwise, with $0<b<a<1$. In this case the lower bound becomes $c(\theta)=\frac{m(m-1)(a-b)}{2\klber(b,a)}$.

\begin{figure}
\begin{center}
\subfigure[]{
\includegraphics[scale=.15]{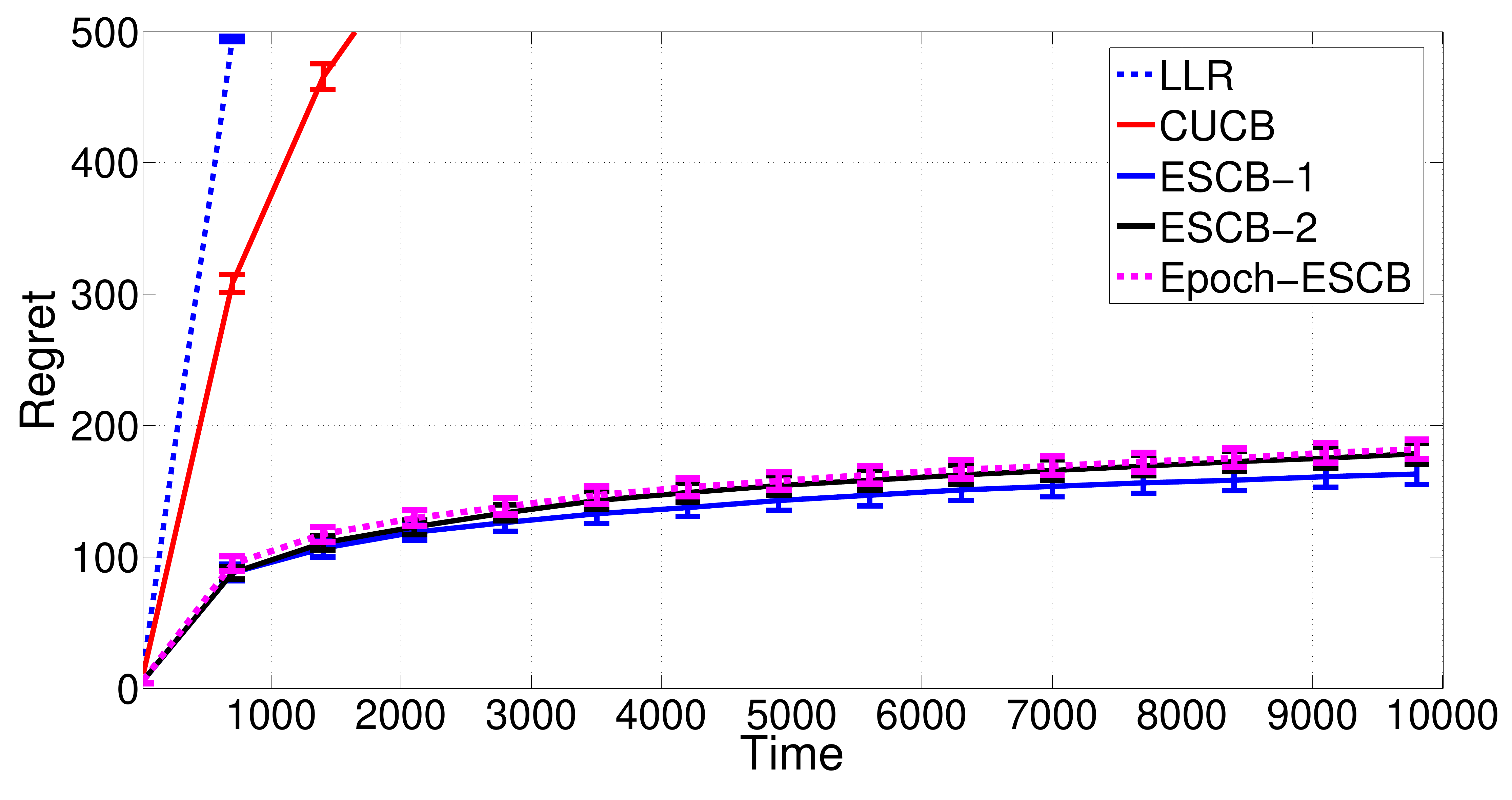}}
\subfigure[]{
\includegraphics[scale=.15]{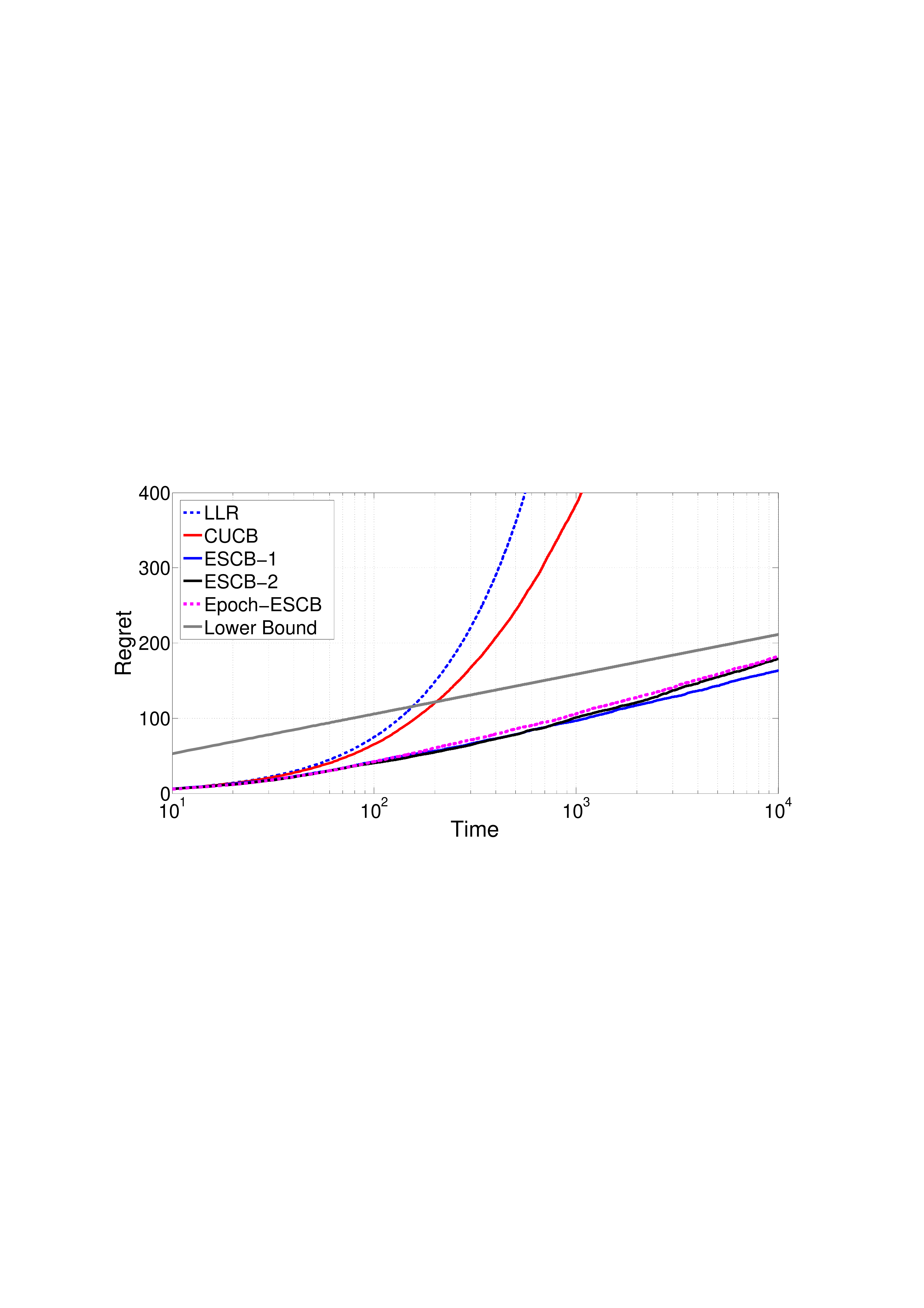}}
\end{center}
\caption{Regret of various algorithms for matchings with $a=0.7$ and $b=0.5$.}
\label{fig:matching_5by5_Expr3}
\end{figure}

Figure \ref{fig:matching_5by5_Expr3}(a)-(b) depicts the regret of various algorithms for the case of $a=0.7$ and $b=0.5$. The curves in Figure \ref{fig:matching_5by5_Expr3}(a) are shown with a 95\% confidence interval. We observe that \algos\textsc{-1} has the lowest regret. Moreover, \algos\textsc{-2} significantly outperforms \textsc{CUCB} and \textsc{LLR}, and is close to \algos\textsc{-1}. Moreover, we observe that the regret of \textsc{Epoch-}\algos attains is quite close to that of \algos-2.

\begin{figure}
\begin{center}
\subfigure[]{
\includegraphics[scale=.15]{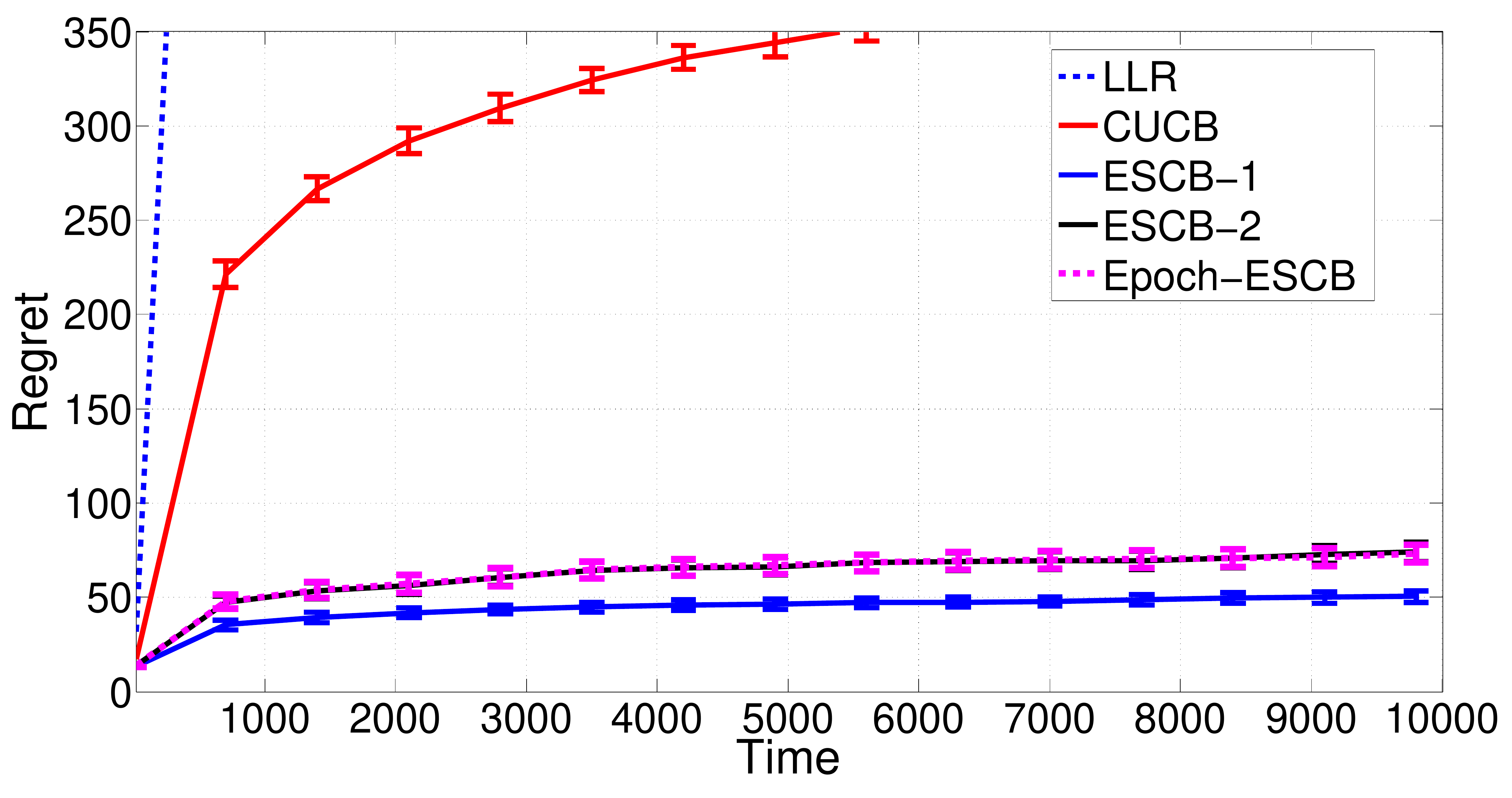}}
\subfigure[]{
\includegraphics[scale=.15]{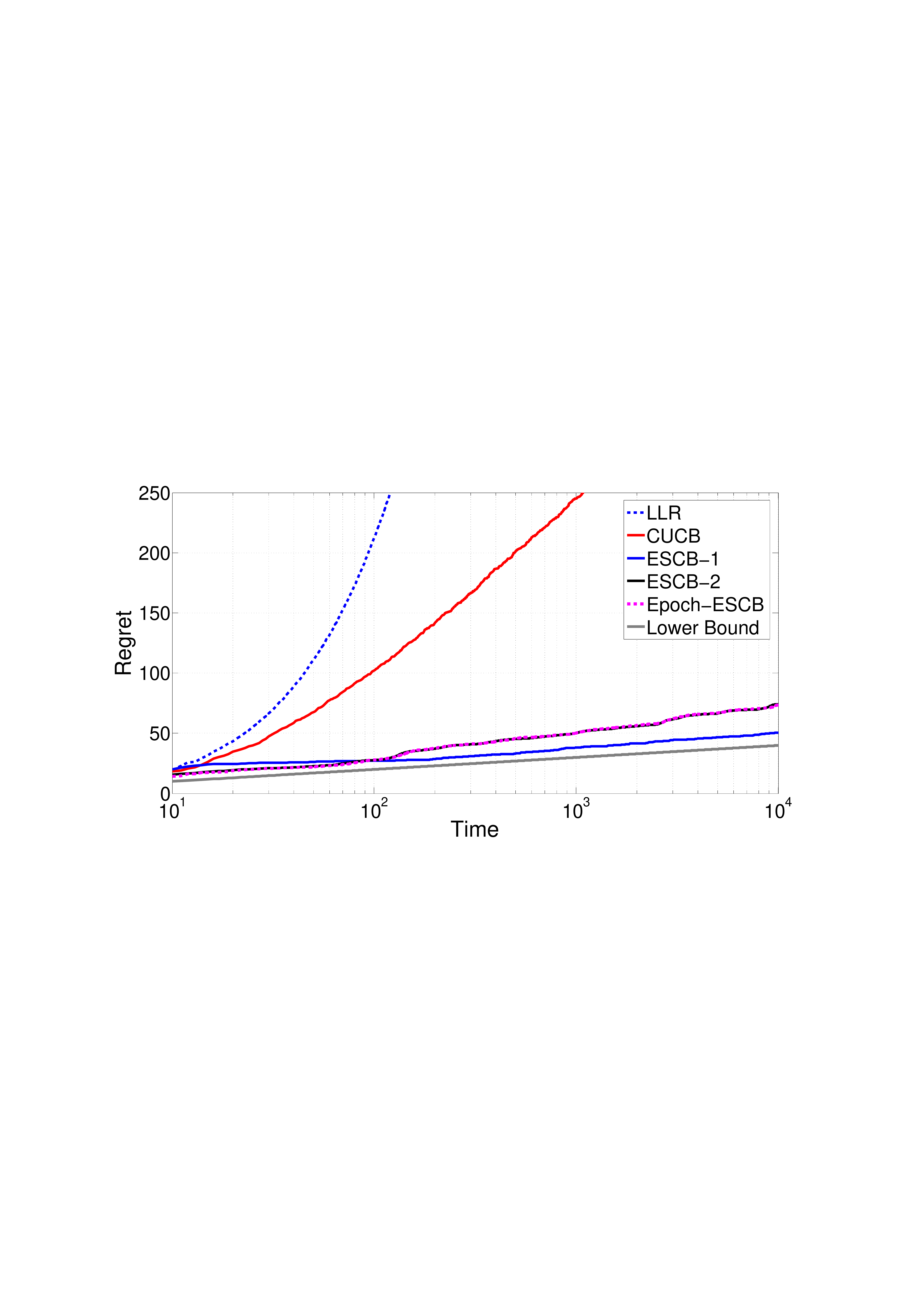}}
\end{center}
\caption{Regret of various algorithms for matchings with $a=0.95$ and $b=0.3$.}
\label{fig:matching_5by5_Expr2}
\end{figure}

Figures \ref{fig:matching_5by5_Expr2}(a)-(b) presents the regret of various algorithms for the case of $a=0.95$ and $b=0.3$.
The difference compared to the former case is that \algos\textsc{-1} significantly outperforms \algos\textsc{-2}. The reason is that in the former case, mean rewards of the most of the basic actions were close to 1/2, for which the performance of UCB-type algorithms are closer to their KL-divergence based counterparts. On the other hand, when mean rewards are not close to 1/2, there exists a significant performance gap between \algos\textsc{-1} and \algos\textsc{-2}. Comparing the results with the `lower bound' curve, we highlight that \algos\textsc{-1} gives close-to-optimal performance in both cases. Furthermore, similar to previous experiment, \textsc{Epoch-}\algos attains a regret whose curve is almost indistinguishable from that of \algos-2.

\begin{figure}
\begin{center}
\subfigure[$a=0.7$ and $b=0.5$]{
\includegraphics[height=1.46in]{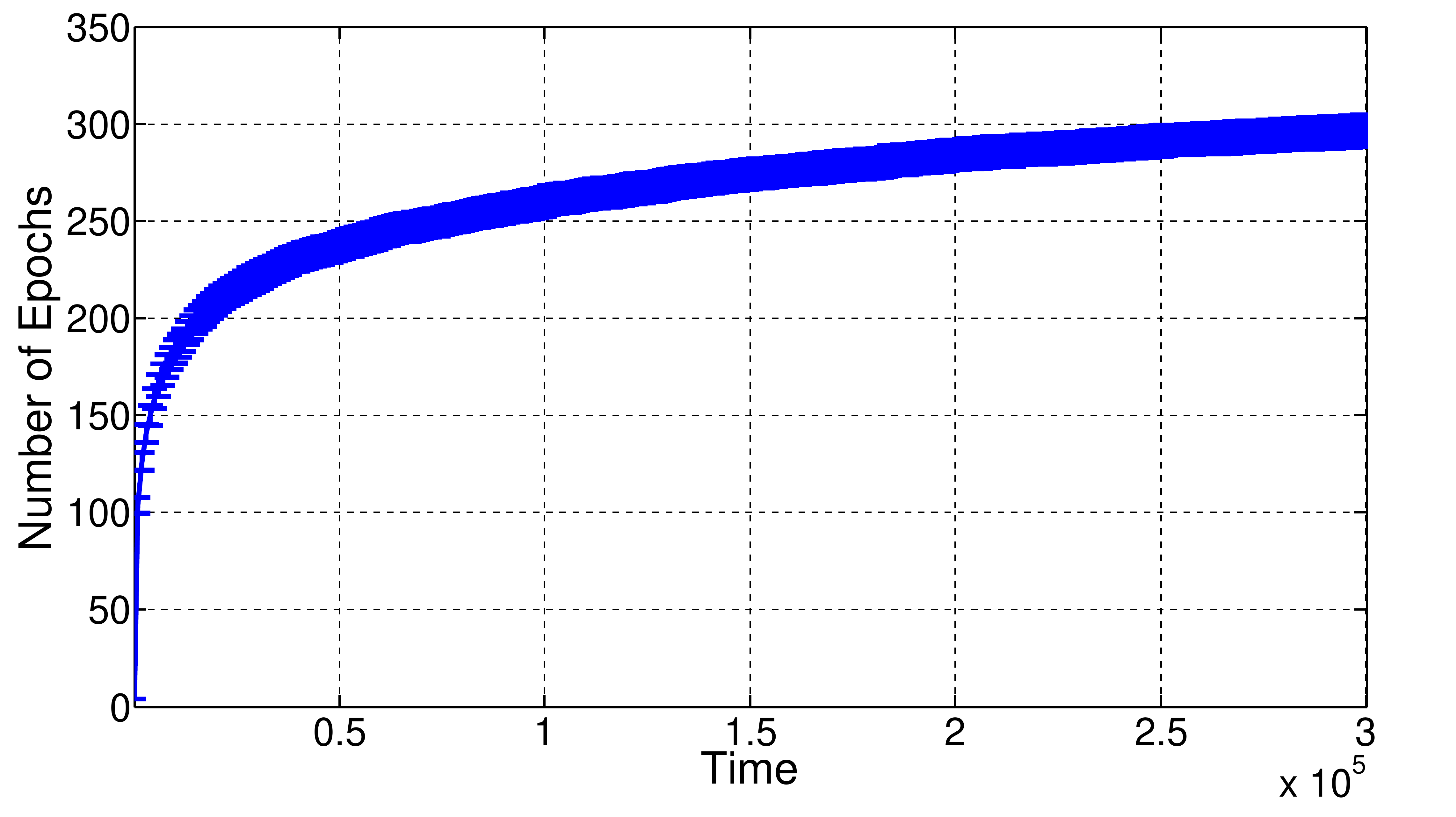}}
\subfigure[$a=0.95$, $b=0.3$]{
\includegraphics[height=1.5in]{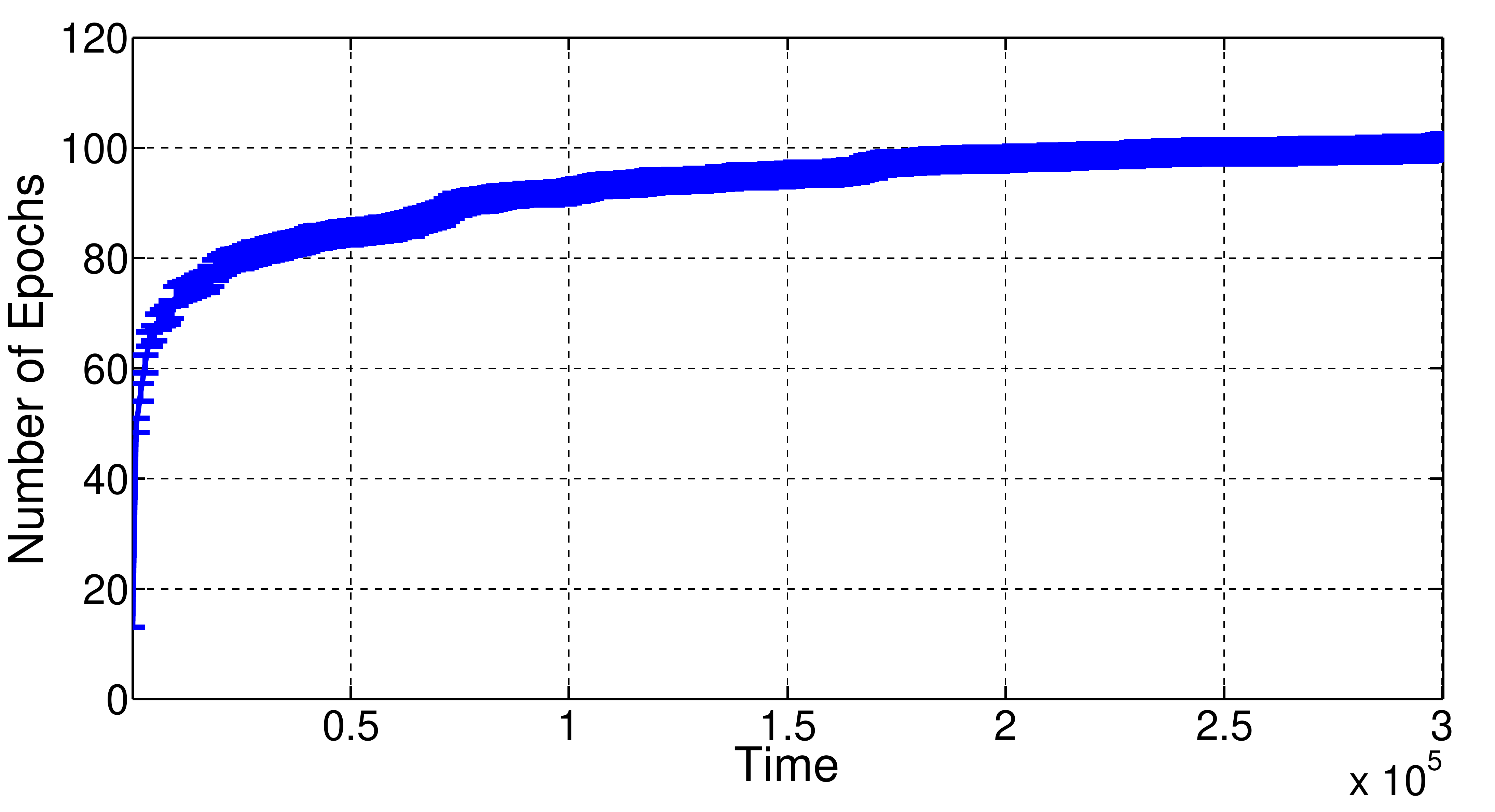}}
\end{center}
\caption{Number of epochs in \textsc{Epoch-}\algos~vs. time for Experiment 1 and 2 (\%95 confidence interval).}
\label{fig:matching_5by5_Epochs}
\end{figure}

The number of epochs in \textsc{Epoch-}\algos~vs. time for the two examples is displayed in Figure \ref{fig:matching_5by5_Epochs}(a)-(b), where the curves are shown with  95\% confidence intervals. We observe that in both cases, the number of epochs grows at a rate proportional to $\log(n)/n$ at round $n$. Since the number of epochs is equal to the number of times the algorithm computes indexes, these curves suggest that index computation after $n$ rounds requires a number of operations that scales as $|\Mcal|\log(n)$.

\subsubsection{Experiment 2: Spanning Trees}
In the second experiment, we consider spanning trees problem described in Section \ref{sec:spanning_tree_prb_1} for the case of $N=5$. In this case, we have $d={5\choose 2}=10$, $m=4$, and $|\Mcal|=5^{3}=125$.

\begin{figure}
\centering
\subfigure[]{
\includegraphics[height=1.4in]{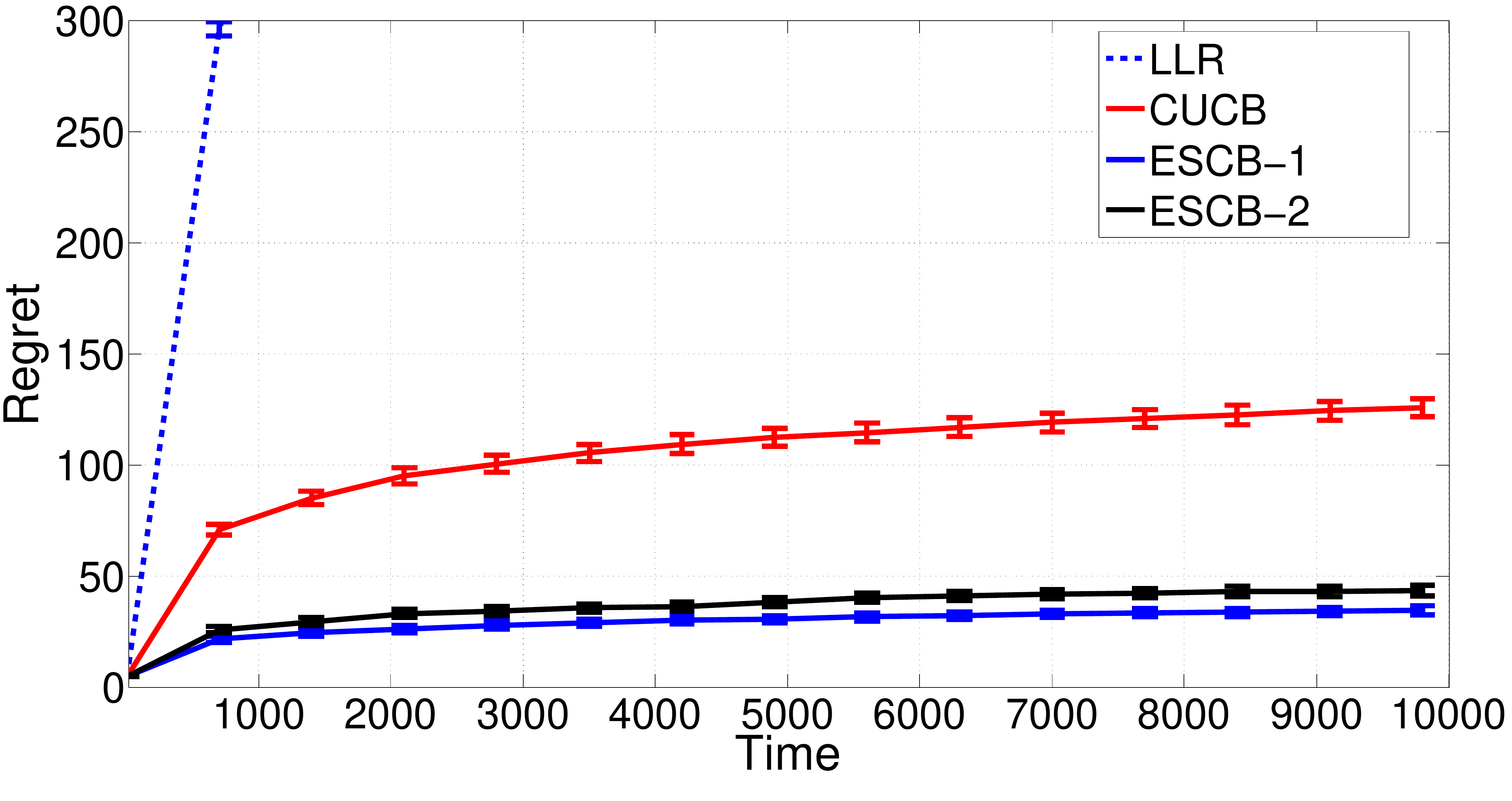}}
\subfigure[]{
\includegraphics[height=1.4in]{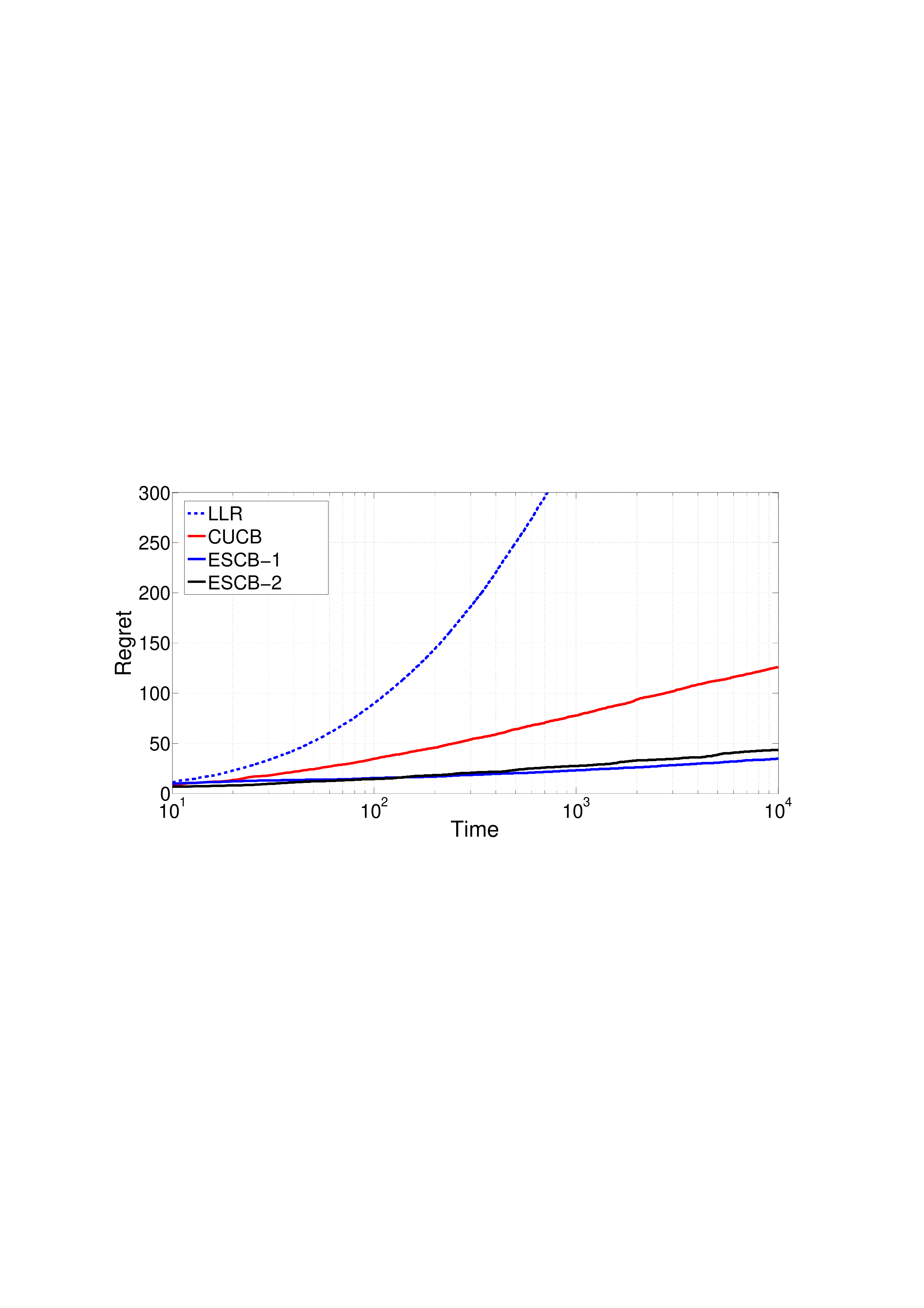}}
\caption{Regret of various algorithms for spanning trees with $N=5$ and  $\Delta_{\min}=0.54$.}
\label{fig:spanning_trees_N5_1}
\end{figure}

Figure \ref{fig:spanning_trees_N5_1} portrays the regret of various algorithms with 95\% confidence intervals, with $\Delta_{\min}=0.54$. Our algorithms significantly outperform \textsc{CUCB} and \textsc{LLR}.

\section{Proofs for Adversarial Combinatorial Bandits}
\label{sec:supp_adversarial}

\subsection{Proof of Theorem 6}

We first prove a simple result:
\begin{lemma}\label{lem:proj}
For all $x\in \mathbb{R}^{d}$, we have
$\Sigma_{n-1}^+\Sigma_{n-1}x=\overline{x}$, where $\overline{x}$ is
the orthogonal projection of $x$ onto $span(\Mcal)$, the linear space spanned by
$\Mcal$.
\end{lemma}

% Insert appropriate proof comment!
\textit{Proof:}
Note that for all $y\in \mathbb{R}^{d}$, if
$\Sigma_{n-1} y=0$, then we have
\begin{align}
\label{eq:0}y^{\top}\Sigma_{n-1} y= \EE\left[ y^{\top}MM^{\top}y\right]  = \EE\left[(y^{\top}M)^2
\right]= 0,
\end{align}
where $M$ has law $p_{n-1}$ such that $\sum_{M}M_ip_{n-1}(M)=q'_{n-1}(i),\;\forall i\in[d]$ and $q'_{n-1}=(1-\gamma)q_{n-1}+\gamma\mu^0$. By definition of $\mu^0$, each $M\in \Mcal$ has a positive probability. Hence, by (\ref{eq:0}), $y^{\top}M=0$ for all $M\in \Mcal$. In particular, we see that the linear application $\Sigma_{n-1}$
restricted to $span(\Mcal)$ is invertible and is zero on $span
(\Mcal)^{\perp}$, hence we have
$\Sigma_{n-1}^+\Sigma_{n-1} x= \overline{x}$.
\ep

%\end{proof}

\begin{lemma}\label{lem:cb2}
We have for any $\eta\le \frac{\gamma\underline{\lambda}}{m^{3/2}}$ and any $q\in \Pcal$,
\begin{align*}
\sum_{n=1}^T q^\top \tilde{X}(n)-\sum_{n=1}^T q_{n-1}^\top \tilde{X}(n)\leq\frac{\eta}{2} \sum_{n=1}^Tq_{n-1}^\top \tilde X^2(n) +\frac{\kl(q,q_0)}{\eta},
\end{align*}
where $\tilde{X}^2(n)$ is the vector that is the coordinate-wise
square of $\tilde{X}(n)$.
\end{lemma}

\textit{Proof:}
We have
\begin{align*}
\kl(q,\tilde{q}_n) - \kl(q,q_{n-1}) =\sum_{i\in[d]} q(i) \log
\frac{q_{n-1}(i)}{\tilde{q}_n(i)}
= -\eta \sum_{i\in[d]}q(i)\tilde{X}_i(n)+\log Z_n,
\end{align*}
with
\begin{align}
\log Z_n &= \log \sum_{i\in[d]} q_{n-1}(i)\exp\left(\eta
  \tilde{X}_i(n)\right)\nonumber\\
\label{eq:Zt_1}
&\leq \log \sum_{i\in[d]} q_{n-1}(i)\left( 1+\eta \tilde{X}_i(n)+\eta^2
    \tilde{X}_i^2(n) \right)\\
\label{eq:Zt_2}
&\leq \eta q_{n-1}^\top \tilde{X}(n) + \eta^2 q_{n-1}^{\top}\tilde{X}^2(n),
\end{align}
where we used $\exp(z)\leq 1+ z+z^2$ for all $|z| \le 1$
in (\ref{eq:Zt_1}) and $\log (1+z)\leq z$ for all $z>-1$ in (\ref{eq:Zt_2}). Later we verify the condition for the former inequality.

Hence we have
\begin{align*}
\kl(q, \tilde{q}_n) - \kl(q, q_{n-1})\leq\eta q_{n-1}^\top \tilde{X}(n)-\eta
q^\top \tilde{X}(n) +\eta^2q_{n-1}^\top\tilde X^2(n).
\end{align*}

Generalized Pythagorean inequality (see Theorem 3.1 in \cite{csishi04}) gives
\begin{align*}
\kl(q,q_n)+\kl(q_n,\tilde{q}_n) \leq \kl(q,\tilde{q}_n).
\end{align*}
Since $\kl(q_n,\tilde{q}_n)\geq 0$, we get
\begin{align*}
\kl(q,q_n)-\kl(q,q_{n-1}) \leq
\eta q_{n-1}^\top \tilde{X}(n) -\eta q^\top \tilde{X}(n) +\eta^2q_{n-1}^\top \tilde X^2(n).
\end{align*}
Finally, summing over $n$ gives
\begin{align*}
\sum_{n=1}^T \left(q^\top \tilde{X}(n)-q_{n-1}^\top \tilde{X}(n)\right)\leq
\eta \sum_{n=1}^Tq_{n-1}^\top \tilde X^2(n) +\frac{\kl(q,q_0)}{\eta}.
\end{align*}

To satisfy the condition for the inequality (\ref{eq:Zt_1}), i.e., $\eta |\tilde X_i(n)|\le 1,\; \forall i\in[d]$, we find the upper bound for $\max_{i\in [d]}  |\tilde{X}_i(n)|$ as follows:
\begin{align*}
\max_{i\in[d]}|\tilde{X}_i(n)| &\le \|\tilde{X}(n)\|_{2} \nonumber\\
            &= \|\Sigma_{n-1}^+ M(n)Y_n\|_{2} \nonumber\\
            &\le m\|\Sigma_{n-1}^+ M(n)\|_{2} \nonumber\\
            &\le m\sqrt{M(n)^\top\Sigma_{n-1}^+\Sigma_{n-1}^+ M(n)} \nonumber\\
            &\le m \|M(n)\|_2\sqrt{\lambda_{\max}\left(\Sigma_{n-1}^+\Sigma_{n-1}^+\right)} \nonumber\\
				&= m^{3/2} \sqrt{\lambda_{\max}\left(\Sigma_{n-1}^+\Sigma_{n-1}^+\right)} \nonumber\\
				&= m^{3/2} \;\lambda_{\max}\left(\Sigma_{n-1}^+\right) \nonumber\\
            &= \frac{m^{3/2}}{\lambda_{\min}\left(\Sigma_{n-1}\right)},\nonumber
\end{align*}
where $\lambda_{\max}(A)$ and $\lambda_{\min}(A)$ respectively denote the maximum and the minimum nonzero eigenvalue of matrix $A$.
Note that $\mu^0$ induces uniform distribution over $\Mcal$. Thus by $q'_{n-1}=(1-\gamma)q_{n-1}+\gamma \mu^0$ we see that $p_{n-1}$ is a mixture of uniform distribution and the distribution induced by $q_{n-1}$.
Note that, we have:
\begin{align*}
\lambda_{\min}\left(\Sigma_{n-1}\right) &= \min_{\|x\|_2=1, x\in span(\Mcal)} x^\top \Sigma_{n-1}x.
\end{align*}
Moreover, we have
\begin{align*}
x^\top \Sigma_{n-1}x &= \EE\left[ x^\top M(n)M(n)^\top x\right] = \EE\left[(M(n)^\top x)^2 \right]\geq \gamma \EE\left[(M^\top x)^2 \right],
\end{align*}
where in the last inequality $M$ has law $\mu^0$. By definition, we have for any $x\in span(\Mcal)$ with $\|x\|_2=1$,
\begin{align*}
\EE\left[(M^\top x)^2 \right] \geq \underline{\lambda},
\end{align*}
so that in the end, we get $\lambda_{\min}(\Sigma_{n-1})\ge \gamma\underline{\lambda}$, and hence $\eta|\tilde{X}_i(n)|\le\frac{\eta m^{3/2}}{\gamma\underline{\lambda}},\; \forall i\in[d]$. Finally, we choose $\eta\le \frac{\gamma\underline{\lambda}}{m^{3/2}}$ to satisfy the condition for the inequality we used in (\ref{eq:Zt_1}).

\ep

We have
\begin{align*}
\EE_n\left[ \tilde{X}(n)\right]= \EE_n\left[
  Y_n\Sigma_{n-1}^{+}M(n)\right]
= \EE_n\left[ \Sigma_{n-1}^{+}M(n)M(n)^{\top} X(n)\right]
= \Sigma_{n-1}^{+}\Sigma_{n-1}X(n)=\overline{X(n)},
\end{align*}
where the last equality follows from Lemma \ref{lem:proj} and
$\overline{X(n)}$ is the orthogonal projection of $X(n)$ onto
$span(\Mcal)$. In particular, for any $mq'\in Co(\Mcal)$, we have
\begin{align*}
\EE_n \left[mq'^\top \tilde{X}(n)\right] = mq'^\top\overline{X(n)}=mq'^\top{X(n)}.
\end{align*}

Moreover, we have:
\begin{align*}
\EE_n\left[ q_{n-1}^\top \tilde{X}^2(n)\right]&= \sum_{i\in [d]} q_{n-1}(i) \EE_n \left[\tilde{X}_i^2(n)\right]\\
&= \sum_{i\in [d]} \frac{q'_{n-1}(i)-\gamma\mu^0(i)}{1-\gamma} \EE_n \left[\tilde{X}_i^2(n)\right]\\
&\le \frac{1}{m(1-\gamma)}\sum_{i\in [d]} mq'_{n-1}(i) \EE_n \left[\tilde{X}_i^2(n)\right]\\
&= \frac{1}{m(1-\gamma)} \EE_n \Bigl[\sum_{i\in [d]}\tilde{M}_i(n)\tilde{X}_i^2(n)\Big],
\end{align*}
where $\tilde{M}(n)$ is a random arm with the same law as $M(n)$ and independent of $M(n)$. Note that $\tilde{M}^2_i(n)=\tilde{M}_i(n)$, so that we have
\begin{align*}
\EE_n \Bigl[\sum_{i\in [d]}\tilde{M}_i(n)\tilde{X}_i^2(n)\Big] &=
\EE_n\left[X(n)^\top M(n) M(n)^\top \Sigma_{n-1}^+ \tilde{M}(n) \tilde{M}(n)^\top \Sigma_{n-1}^+ M(n) M(n)^\top X(n) \right]\\
&\le m^2\EE_n[M(n)^\top \Sigma_{n-1}^+ M(n)],
\end{align*}
where we used the bound $M(n)^\top X(n)\le m$. By \cite[Lemma~15]{cesa2012}, $\EE_n[M(n)^\top \Sigma_{n-1}^+ M(n)]\le d$, so that we have:
$$
\EE_n\left[ q_{n-1}^\top \tilde{X}^2(n)\right]\le \frac{md}{1-\gamma}.
$$

Observe that
\begin{align*}
\EE_n\left[ q^{\star\top} \tilde{X}(n)-q_{n-1}'^{\top} \tilde{X}(n)\right]
&= \EE_n\left[ q^{\star\top} \tilde{X}(n)-(1-\gamma)q_{n-1}^\top \tilde{X}(n)-\gamma\mu^{0\top} \tilde{X}(n)\right]\\
&= \EE_n\left[ q^{\star\top} \tilde{X}(n)-q_{n-1}^\top \tilde{X}(n)\right]+\gamma q_{n-1}^\top X(n)-\gamma\mu^{0\top} X(n)\\
&\le \EE_n\left[ q^{\star\top} \tilde{X}(n)-q_{n-1}^\top \tilde{X}(n)\right]+\gamma q_{n-1}^\top X(n)\\
&\le \EE_n\left[ q^{\star\top} \tilde{X}(n)-q_{n-1}^\top \tilde{X}(n)\right]+\gamma.
\end{align*}

Using Lemma \ref{lem:cb2} and the above bounds, we get with $mq^\star$ the optimal arm, i.e. $q^\star(i)=\frac{1}{m}$ iff $M^\star_i=1$,
\begin{align*}
R^{\textsc{CombEXP}}(T) &= \EE\Bigl[ \sum_{n=1}^T mq^{\star\top} \tilde{X}(n)-\sum_{n=1}^T mq_{n-1}'^{\top} \tilde{X}(n)\Big]\\
&\le \EE\Bigl[ \sum_{n=1}^T mq^{\star\top} \tilde{X}(n)-\sum_{n=1}^T mq_{n-1}^\top \tilde{X}(n)\Big]+m\gamma T\\
&\le \frac{\eta m^2dT}{1-\gamma}+\frac{m\log \mu^{-1}_{\min}}{\eta}+m\gamma T,
\end{align*}
since
\begin{align*}
\kl(q^\star,q_0) = -\frac{1}{m}\sum_{i\in M^\star}\log m\mu^{0}_{i}\leq \log \mu^{-1}_{\min}.
\end{align*}

Choosing $\eta=\gamma C$ with $C=\frac{\underline{\lambda}}{m^{3/2}}$ gives
\begin{align*}
R^{\textsc{CombEXP}}(T)&\le \frac{\gamma Cm^2dT}{1-\gamma}+\frac{m\log \mu^{-1}_{\min}}{\gamma C}+m\gamma T\\
        &= \frac{Cm^2d + m -m\gamma}{1-\gamma}\gamma T+\frac{m\log \mu^{-1}_{\min}}{\gamma C}\\
        &\le \frac{(Cm^2d+m)\gamma T}{1-\gamma}+\frac{m\log \mu^{-1}_{\min}}{\gamma C}.
\end{align*}
The proof is completed by setting
$
\gamma = \frac{\sqrt{m\log \mu^{-1}_{\min}}}{\sqrt{m\log \mu^{-1}_{\min}}+\sqrt{C(Cm^2d+m)T}}.
$
\ep

\subsection{Proof of Proposition 1}

We first provide a simple result:

\begin{lemma}\label{lem:convexity_kl}
The KL-divergence $z\mapsto \kl(z,q)$ is 1-strongly convex with respect to the $\|\cdot\|_1$ norm.
\end{lemma}

\bp
To prove the lemma, it suffices to show that for any $x,y\in \Pcal$:
\begin{align*}
(\nabla \kl(x,q)-\nabla \kl(y,q))^\top(x-y) \ge \|x-y\|_1^2.
\end{align*}
We have
\begin{align*}
(\nabla \kl(x,q)-\nabla \kl(y,q))^\top(x-y)&=\sum_{i\in [d]} \Bigl(1+\log \frac{x(i)}{q(i)}-1-\log \frac{y(i)}{q(i)}\Big)(x(i)-y(i)) \\ &=\sum_{i\in [d]} (1+\log x(i)-1-\log y(i))(x(i)-y(i))\\
&=\Bigl(\nabla \sum_{i\in [d]} x(i)\log x(i)-\nabla \sum_{i\in [d]} y(i)\log y(i)\Big)^\top (x-y)\\
&\ge \|x-y\|_1^2,
\end{align*}
where the last inequality follows from strong convexity of the entropy function $z\mapsto \sum_{i\in [d]} z_i\log z_i$ with respect to the $\|\cdot\|_1$ norm \cite[Proposition~5.1]{beck2003mirror}.
\ep

Recall that $u_n=\arg\min_{p\in\Pcal} \kl(p,\tilde q_n)$ and that $q_n$ is an $\epsilon_n$-optimal solution for the projection step, that is
\begin{align*}
\kl(u_n,\tilde q_n)\ge \kl(q_n,\tilde q_n)-\epsilon_n.
\end{align*}
By Lemma \ref{lem:convexity_kl}, we have
\begin{align*}
\kl(q_n,\tilde q_n)-\kl(u_n,\tilde q_n)&\ge (q_n-u_n)^\top\nabla\kl(u_n,\tilde q_n)+ \frac{1}{2}\|q_n-u_n\|_1^2\ge \frac{1}{2}\|q_n-u_n\|_1^2,
\end{align*}
where we used $(q_n-u_n)^\top\nabla\kl(u_n,\tilde q_n)\ge 0$ due to first-order optimality condition for $u_n$.
Hence $\kl(q_n,\tilde q_n)-\kl(u_n,\tilde q_n)\le \epsilon_n$ implies that $\|q_n-u_n\|_\infty\le \|q_n-u_n\|_1\le \sqrt{2\epsilon_n}$.

Consider $q^\star$, the distribution over $\Pcal$ for the optimal arm, i.e. $q^\star(i)=\frac{1}{m}$ iff $M^\star_i=1$. Recall that from proof of Lemma \ref{lem:cb2}, for $q=q^\star$ we have
\begin{align}
\label{eq:lem_cb2_ineq}
\kl(q^\star, \tilde{q}_n) - \kl(q^\star, q_{n-1})\leq\eta q_{n-1}^\top \tilde{X}(n)-\eta
{q^\star}^\top \tilde{X}(n) +\eta^2q_{n-1}^\top\tilde X^2(n).
\end{align}
Generalized Pythagorean Inequality (see Theorem 3.1 in \cite{csishi04}) gives
\begin{align}
\label{eq:GPI_u_n}
\kl(q^\star,\tilde q_n) &\ge \kl(q^\star,u_n)+\kl(u_n,\tilde q_n).
\end{align}
Let $\underline{q}_n=\min_{i} q_n(i)$. Observe that
\begin{align*}
\kl(q^\star,u_n)&=\sum_{i\in [d]} q^\star(i)\log \frac{q^\star(i)}{u_n(i)}= -\frac{1}{m} \sum_{i\in M^\star} \log mu_n(i)\\
&\ge -\frac{1}{m} \sum_{i\in M^\star} \log m(q_n(i)+\sqrt{2\epsilon_n}) \ge -\frac{1}{m} \sum_{i\in M^\star} \Bigl(\log mq_n(i)+ \frac{\sqrt{2\epsilon_n}}{\underline{q}_n}\Big)\\
&\ge -\frac{\sqrt{2\epsilon_n}}{\underline{q}_n} -\frac{1}{m} \sum_{i\in M^\star} \log mq_n(i)=
-\frac{\sqrt{2\epsilon_n}}{\underline{q}_n} + \kl(q^\star,q_n),
\end{align*}
%where we used $\log(1+z)\le z$ for all $z>-1$ in the second line.
Plugging this into (\ref{eq:GPI_u_n}), we get
\begin{align*}
\kl(q^\star,\tilde q_n) &\ge \kl(q^\star,q_n)-\frac{\sqrt{2\epsilon_n}}{\underline{q}_n}+\kl(u_n,\tilde q_n)\ge \kl(q^\star,q_n)-\frac{\sqrt{2\epsilon_n}}{\underline{q}_n}.
\end{align*}
Putting this together with (\ref{eq:lem_cb2_ineq}) yields
\begin{align*}
\kl(q^\star,q_n)-\kl(q^\star,q_{n-1}) \leq
\eta q_{n-1}^\top \tilde{X}(n) -\eta {q^\star}^\top \tilde{X}(n) +\eta^2q_{n-1}^\top \tilde X^2(n)+\frac{\sqrt{2\epsilon_n}}{\underline{q}_n}.
\end{align*}
Finally, summing over $n$ gives
\begin{align*}
\sum_{n=1}^T \left({q^\star}^\top \tilde{X}(n)-q_{n-1}^\top \tilde{X}(n)\right)&\leq
\eta \sum_{n=1}^Tq_{n-1}^\top \tilde X^2(n) +\frac{\kl(q^\star,q_0)}{\eta}+\frac{1}{\eta}\sum_{n=1}^T \frac{\sqrt{2\epsilon_n}}{\underline{q}_n}.
\end{align*}

Defining
$$
\epsilon_n=\frac{\left(\underline{q}_n\log\mu_{\min}^{-1}\right)^2}{32n^{2}\log^{3}(n)},\;\; \forall n\ge 1,
$$
and recalling that $\kl(q^\star,q_0) \le \log \mu^{-1}_{\min}$, we get
\begin{align*}
\sum_{n=1}^T \left({q^\star}^\top \tilde{X}(n)-q_{n-1}^\top \tilde{X}(n)\right)
&\leq
\eta \sum_{n=1}^Tq_{n-1}^\top \tilde X^2(n) +\frac{\log \mu^{-1}_{\min}}{\eta}+\frac{\log \mu^{-1}_{\min}}{\eta}\sum_{n=1}^T \sqrt{\frac{2}{32n^2\log^3(n+1)}}\\
&\leq
\eta \sum_{n=1}^Tq_{n-1}^\top \tilde X^2(n) +\frac{2\log \mu^{-1}_{\min}}{\eta},
\end{align*}
where we used the fact $\sum_{n\ge 1} n^{-1}(\log(n+1))^{-3/2}\le 4$.
We remark that by the properties of KL divergence and since $q'_{n-1}\ge \gamma\mu^0>0$, we have $\underline{q}_n>0$ at every round $n$, so that $\epsilon_n>0$ at every round $n$.

Using the above result and following the same lines as in the proof of Theorem 6, we have
\begin{align*}
R^{\textsc{CombEXP}}(T) &\le \frac{\eta m^2dT}{1-\gamma}+\frac{2m\log \mu^{-1}_{\min}}{\eta}+m\gamma T.
\end{align*}
Choosing $\eta=\gamma C$ with $C=\frac{\underline{\lambda}}{m^{3/2}}$ gives
\begin{align*}
R^{\textsc{CombEXP}}(T) &\le \frac{(Cm^2d+m)\gamma T}{1-\gamma}+\frac{2m\log \mu^{-1}_{\min}}{\gamma C}.
\end{align*}
The proof is completed by setting
$
\gamma = \frac{\sqrt{2m\log \mu^{-1}_{\min}}}{\sqrt{2m\log \mu^{-1}_{\min}}+\sqrt{C(Cm^2d+m)T}}.
$
\ep

\subsection{Proof of Theorem 7}

We calculate the time complexity of the various steps of \textsc{CombEXP} at round $n\ge 1$.

\begin{enumerate}
\item[(i)] \underline{Mixing:} This step requires $\Ocal(d)$ time.

\item[(ii)] \underline{Decomposition:} Using the algorithm of \cite{sherali1987constructive}, the vector $mq'_{n-1}$ may be  represented as a convex combination of at most $d+1$ arms in $\Ocal(d^4)$ time, so that $p_{n-1}$ may have at most $d+1$ non-zero elements (observe that the existence of such a representation follows from Carath\'eodory Theorem).

\item[(iii)] \underline{Sampling:} This step takes $\Ocal(d)$ time since $p_{n-1}$ has at most $d+1$ non-zero elements.

\item[(iv)] \underline{Estimation:} The construction of matrix $\Sigma_{n-1}$ is done in time $\Ocal(d^2)$ since $p_{n}$ has at most $d+1$ non-zero elements and $MM^\top$ is formed in $\Ocal(d)$ time. Computing the pseudo-inverse of $\Sigma_{n-1}$ costs $\Ocal(d^3)$.

\item[(v)] \underline{Update:} This step requires $\Ocal(d)$ time.

\item[(vi)] \underline{Projection:} The projection step is equivalent to solving a convex program up to accuracy $\epsilon_n=\Ocal(n^{-2}\log^{-3}(n))$. We use the Interior-Point Method (Barrier method). The total number of Newton iterations to achieve accuracy $\epsilon_n$ is $\Ocal(\sqrt{s}\log(s/\epsilon_n))$ \cite[Ch.~11]{boyd2004convex}. Moreover, the cost of each iteration is $\Ocal((d+c)^3)$ \cite[Ch.~10]{boyd2004convex}, so that the total cost of this step becomes $\Ocal(\sqrt{s}(c+d)^3\log(s/\epsilon_n))$. Plugging $\epsilon_n=\Ocal(n^{-2}\log^{-3}(n))$ and noting that $\Ocal(\sum_{n=1}^T \log(s/\epsilon_n))=\Ocal(T\log(T))$, the cost of this step is $\Ocal(\sqrt{s}(c+d)^3T\log(T))$.
\end{enumerate}

Hence the total time complexity after $T$ rounds is $\Ocal(T[\sqrt{s}(c+d)^3\log(T)+d^4])$, which completes the proof.
\ep

\subsection{Implementation: The Case of Graph Coloring}
In this subsection, we present an iterative algorithm for the projection step of \textsc{CombEXP}, for the graph coloring problem described next.

Consider a graph $G=(V,E)$ consisting of $m$ nodes indexed by $i\in[m]$. Each node can use one of the
$c\ge m$ available colors indexed by $j\in[c]$. A feasible coloring is represented by a matrix $M\in\{0,1\}^{m\times c}$, where
$M_{ij} = 1$ if and only if node $i$ is assigned color $j$.
Coloring $M$ is feasible if (i) for all $i$, node $i$ uses at most one color, i.e., $\sum_{j\in [c]}M_{ij}\in \{ 0,1\}$; (ii) neighboring nodes are assigned different colors, i.e., for all $i,i'\in [m]$, $(i,i')\in E$ implies for all $j\in [c]$, $M_{ij}M_{i'j}=0$. In the following we denote by ${\cal K}=\{ {\cal K}_{\ell}, \ell\in [k]\}$ the set of maximal cliques of the graph $G$. We also introduce $K_{\ell i}\in \{0,1\}$ such that $K_{\ell i}=1$ if and only if node $i$ belongs to the maximal clique ${\cal K}_\ell$.

There is a specific case where our algorithm can be efficiently
implementable: when the convex hull $Co(\Mcal)$ can be captured by
polynomial in $m$ many constraints. Note that this cannot be ensured
unless restrictive assumptions are made on the graph $G$
since there are up to $3^{m/3}$ maximal cliques in a graph with $m$
vertices \cite{momo65}. There are families of graphs in which the number of cliques is polynomially bounded. These families include chordal graphs, complete graphs, triangle-free graphs, interval graphs, and planar graphs. Note however, that a limited number of cliques does not ensure a priori that $Co(\Mcal)$ can be captured by a limited number of constraints. To the best of our knowledge, this problem is open and only particular cases have been solved as for the stable set polytope (corresponding to the case $c=2$, $X_{i1}=1$ and $X_{i2}=0$ with our notation) \cite{sch04}.

For the coloring problem described above we have
\begin{align}
\label{eq:CoM}
Co(\Mcal) = Co\{\forall i, \sum_{j\in [c]} M_{ij}\leq 1,\quad \forall \ell,j, \sum_{i\in [m]} K_{\ell i}M_{ij}\leq 1\}.
\end{align}
Note that in the special case where $G$ is the complete graph, such a representation becomes
\begin{align*}
Co(\Mcal)=Co\{\sum_{j\in [c]} M_{ij}\leq 1,\quad \forall i,\:\sum_{i\in [m]} M_{ij}\leq 1,\quad \forall j\}.
\end{align*}
We now give an algorithm for the projection a distribution $p$ onto $\Pcal$ using KL divergence. Since $\Pcal$ is a scaled version of $Co(\Mcal)$, we give an algorithm for the projection of $mp$ onto $Co(\Mcal)$ given by (\ref{eq:CoM}).

Set $\lambda_i(0)=\mu_j(0)=0$ for all $i,j$ and then define for $t\geq 0$,
\begin{align}
\label{eq:iter1}\forall i\in [m],\:\lambda_i(t+1) =& \log \Bigl(\sum_jmp_{ij}e^{-\mu_j(t)} \Big)\\
\label{eq:iter2}\forall j\in[c],\:\mu_j(t+1) =&\max_{\ell}\log\Bigl( \sum_i K_{i\ell}mp_{ij}e^{-\lambda_i(t+1)}\Big).
\end{align}
We can show that
\begin{proposition}
Let $p^\star_{ij}=\lim_{t\to \infty} p_{ij}e^{-\lambda_i(t)-\mu_j(t)}$. Then $mp^\star$ is the projection of $mp$ onto $Co(\Mcal)$ using the KL divergence.
\end{proposition}
Although this algorithm is shown to converge, we must stress that the step (\ref{eq:iter2}) might be expensive as the number of distinct values of $\ell$ might be exponential in $m$. When $G$ is a complete graph, this step is easy and our algorithm reduces to Sinkhorn's algorithm (see \cite{helmbold2009} for a discussion).

\textit{Proof:}
First note that the definition of projection can be extended to non-negative vectors thanks to the relation
$$
\kl(p^\star,q)=\min_{p\in \Xi}\kl(p,q).
$$
More precisely, given an alphabet $A$ and a vector $q\in \mathbb{R}_+^A$, we have for any probability vector $p\in \mathbb{R}_+^A$
\begin{align*}
\sum_{a\in A}p(a)\log \frac{p(a)}{q(a)} \geq \sum_ap(a)\log \frac{\sum_ap(a)}{\sum_aq(a)}
=\log \frac{1}{\|q\|_1},
\end{align*}
thanks to the log-sum inequality. Hence we see that $p^\star(a)=\frac{q(a)}{\|q\|_1}$ is the projection of $q$ onto the simplex of $\mathbb{R}_+^A$.

Now define $\mathcal{A}_i=Co\{M_{ij}, \sum_j M_{ij}\le 1\}$ and $\mathcal{B}_{\ell j}=Co\{M_{ij}, \sum_i K_{\ell i}M_{ij}\leq 1\}$. Hence $\bigcap_i\mathcal{A}_i\bigcap\bigcap_{\ell j}\mathcal{B}_{\ell j} = Co(\Mcal)$. By the argument described above, iteration (\ref{eq:iter1}) (resp. (\ref{eq:iter2})) corresponds to the projection onto $\mathcal{A}_i$ (resp. $\bigcap_\ell\mathcal{B}_{\ell j}$) and the proposition follows from Theorem 5.1 in \cite{csishi04}.
\ep

\subsection{Examples}
In this subsection, we compare the performance of \textsc{CombEXP} against state-of-the-art algorithms (refer to Table 2 for the summary of regret of various algorithms).

\subsubsection{$m$-sets}
In this case, $\Mcal$ is the set of all $d$-dimensional binary vectors with $m$ ones. We have
$$
\mu_{\min}=\min_i \frac{1}{{d\choose m}}\sum_M M_i=\frac{{{d-1}\choose {m-1}}}{{d\choose m}}=\frac{m}{d}.
$$
Moreover, according to \cite[Proposition 12]{cesa2012}, we have
$
\underline{\lambda}=\frac{m(d-m)}{d(d-1)}.
$
When $m=o(d)$, the regret of \textsc{CombEXP} becomes $O(\sqrt{m^{3}d T\log(d/m)})$, namely it has the same performance as \textsc{ComBand} and \textsc{EXP2 with John's Exploration}.

\subsubsection{Matching}
Let $\Mcal$ be the set of perfect matchings in $\Kcal_{m,m}$, where we have $d=m^2$ and $|\Mcal|=m!$. We have
$$
\mu_{\min}=\min_i \frac{1}{m!}\sum_{M} M_i=\frac{(m-1)!}{m!}=\frac{1}{m},
$$
Furthermore, from \cite[Proposition 4]{cesa2012} we have that $\underline{\lambda}=\frac{1}{m-1}$, thus giving $R^\textsc{CombEXP}(T)=O(\sqrt{m^{5} T\log(m)})$, which is the same as the regret of \textsc{ComBand} and \textsc{EXP2 with John's Exploration} in this case.

\subsubsection{Spanning Trees}
In our next example, we assume that $\Mcal$ is the set of spanning trees in the complete graph $\Kcal_{N}$. In this case, we have $d={N\choose 2}$, $m=N-1$, and by Cayley's formula $\Mcal$ has $N^{N-2}$ elements.
Observe that
$$
\mu_{\min}=\min_i \frac{1}{N^{N-2}}\sum_{M} M_i=\frac{(N-1)^{N-3}}{N^{N-2}},
$$
which gives for $N\ge 2$
\begin{align*}
\log\mu_{\min}^{-1}&=\log\left(\frac{N^{N-2}}{(N-1)^{N-3}}\right)\\
					&=(N-3)\log\left(\frac{N}{N-1}\right)+\log N\\
					&\le(N-3)\log 2 + \log(N)\le 2N.
\end{align*}
From \cite[Corollary 7]{cesa2012}, we also get $\underline{\lambda}\ge  \frac{1}{N}-\frac{17}{4N^2}$.
For $N\ge 6$, the regret of \textsc{ComBand} takes the form $O(\sqrt{N^5T\log(N)})$ since $\frac{m}{d\underline{\lambda}}<7$ when $N\ge 6$. Further, \textsc{EXP2 with John's Exploration} attains the same regret. On the other hand, we get
$$
R^\textsc{CombEXP}(T)=O(\sqrt{N^{5}T\log(N)}), \quad N\ge 6,
$$
and therefore it gives the same regret as \textsc{ComBand} and \textsc{EXP2 with John's Exploration}.

\subsubsection{Cut sets}
Consider the case where $\Mcal$ is the set of balanced cuts of the complete graph $\Kcal_{2N}$, where a balanced cut is defined as the set of edges between a set of $N$ vertices and its complement. It is easy to verify that $d={{2N}\choose 2}$ and $m=N^2$. Moreover, $\Mcal$ has ${2N\choose N}$ balanced cuts and hence
$$
\mu_{\min}=\min_i \frac{1}{{2N\choose N}}\sum_{M} M_i=\frac{{2N-2\choose N-1}}{{2N\choose N}}=\frac{N}{4N-2},
$$
Moreover, by \cite[Proposition 9]{cesa2012}, we have
$$
\underline{\lambda}=\frac{1}{4}+\frac{8N-7}{4(2N-1)(2N-3)},\quad N\ge 2,
$$
and consequently, the regret of \textsc{CombEXP} becomes $O(N^{4}\sqrt{T})$ for $N\ge 2$, which is the same as that of  \textsc{ComBand} and \textsc{EXP2 with John's Exploration}.

\end{document}